\documentclass[runningheads]{llncs}

\usepackage[T1]{fontenc}

\usepackage{graphicx}

\usepackage{color}
\usepackage[table]{xcolor}
\usepackage{url}

\usepackage{todonotes}
\usepackage{amssymb}
\usepackage{amsmath}
\usepackage{svg}
\usepackage{tikz}
\usepackage{pgfplots}
\pgfplotsset{compat=1.18}
\usepackage{multicol}
\usepackage{multirow}
\usepackage{paralist}
\usepackage{subcaption}
\usepackage{algorithm}
\usepackage[noend]{algpseudocode}
\usepackage{hyperref}
\usepackage{wrapfig}

\newcommand{\enc}{\pi}
\newcommand{\enci}{\rho}

\usepackage[acronym,nonumberlist]{glossaries}

\newacronym{LDA}{LDA}{linear discriminant analysis}
\newacronym{LR}{LR}{logistic regression}
\newacronym{RNN}{RNN}{recurrent neural network}

\newacronym[plural=DFAs,firstplural=deterministic finite automata (DFA)]{DFA}{DFA}{deterministic finite automaton}

\begin{document}
\title{On the Relationship Between RNN Hidden State Vectors and Semantic Ground Truth}
% \title{Hidden-State Vectors of Recurrent Neural Networks: Do They Cluster?}
\titlerunning{Clustering of RNN Hidden State Vectors}
%
%\titlerunning{s
%
% \author{
% Edi Mu\v{s}kardin\inst{1,2}\orcidID{0000-0001-8089-5024} \and
% Martin Tappler\inst{2,1}\orcidID{0000-0002-4193-5609} \and
% Ingo Pill\inst{1}\orcidID{0000-0002-8420-6377} \and
% Bernhard~K.~Aichernig\inst{2}\orcidID{0000-0002-3484-5584} \and
% Thomas Pock\inst{3,1}\orcidID{0000-0001-6120-1058}
% }
\author{
Edi Mu\v{s}kardin\inst{1,2} \and
Martin Tappler\inst{2,1} \and
Ingo Pill\inst{1} \and \\
Bernhard~K.~Aichernig\inst{2} \and
Thomas Pock\inst{3,1}
}

\institute{
Silicon Austria Labs, TU Graz - SAL DES Lab, Graz, Austria
\and
Graz University of Technology, Institute of Software Technology, Graz, Austria
\and
Graz University of Technology, Institute of Computer Graphics and Vision, Graz, Austria
}

\authorrunning{Mu\v{s}kardin et al.}

\maketitle              % typeset the header of the contribution
\begin{abstract}
We examine the assumption that the hidden-state vectors of recurrent neural networks (RNNs) tend to form clusters of semantically similar vectors, which we dub the \emph{clustering hypothesis}. While this hypothesis has been assumed in the analysis of RNNs in recent years, its validity has not been studied thoroughly on modern neural network architectures. We examine the clustering hypothesis in the context of RNNs that were trained to recognize regular languages. This enables us to draw on perfect ground-truth automata in our evaluation, against which we can compare the RNN's accuracy and the distribution of the hidden-state vectors.

We start with examining the (piecewise linear) separability of an RNN's hidden-state vectors into semantically different classes. We continue the analysis by computing clusters over the hidden-state vector space with multiple state-of-the-art unsupervised clustering approaches. We formally analyze the accuracy of computed clustering functions and the validity of the clustering hypothesis by determining whether clusters group semantically similar vectors to the same state in the ground-truth model.
 
Our evaluation supports the validity of the clustering hypothesis in the majority of examined cases. We observed that the hidden-state vectors of well-trained RNNs are separable, and that the unsupervised clustering techniques succeed in finding clusters of similar state vectors.

\end{abstract}

\section{Introduction}\label{sec:intro}

% BKA
In recent years we have seen significant advancements in the analysis and verification of artificial neural networks (ANNs) ~\cite{DBLP:conf/cav/HuangKWW17,DBLP:journals/corr/abs-1710-00486,DBLP:conf/kbse/SunWRHKK18}. The motivation has been to establish trust in the growing landscape of machine-learned intelligent systems. Most of the work considered feed-forward ANNs that implement mathematical functions without internal states. In this work we shift the focus to recurrent neural networks (RNNs) that model processes with internal states. RNNs have shown impressive results on high-dimensional data, like natural-language, but their internal decision-making progress is hard to interpret. 

Consequently, many researchers have turned to the internals of RNNs in hopes of understanding the RNN decision-making process~\cite{DBLP:journals/nn/OmlinG96}. 
An important, yet scarcely examined, hypothesis in this field postulates that the hidden state vectors visited by a trained RNN while processing data form clusters. The states in a cluster are assumed to be semantically similar.
For example, an RNN processing two semantically similar, but syntactically
different natural-language sentences is assumed to traverse states that belong to the same state clusters.

Hence, the ``\textit{clustering hypothesis}'' provides powerful
analysis capabilities. It facilitates the creation
of equivalence classes in the internal state space of RNNs by clustering. Put differently, it enables the discretization of RNN hidden state vectors to a finite number of equivalence classes.
These classes enable further analyses, e.g., through the extraction of finite-state machines (FSMs) with states corresponding to clusters. Unlike RNNs, FSMs are well-understood and amenable to model-based reasoning. In recent years, the clustering hypothesis was the basis of several well received RNN analyses methods. For example, Dong et al.~\cite{DBLP:conf/kbse/DongWSZWDDW20} relied on the clustering hypothesis to serve as an abstraction mechanism for models which enable identifications of adversarial attacks, while Michalenko et al.~\cite{DBLP:conf/iclr/MichalenkoSVBCP19} learned functions from hidden states to (sets of) DFA states. While both approaches relied on some form of the clustering hypothesis, they also encountered, and subsequently coped with inaccuracies of computed clustering function. For example, \cite{DBLP:conf/kbse/DongWSZWDDW20} used probabilistic automata learning to minimize the error introduced by the computed clustering. While their approaches \textit{might imply} the soundness of the clustering hypothesis, the validity of the hypothesis still remains questioned. For example, already in early work Kolen~\cite{DBLP:conf/nips/Kolen93} warned against methods that rely on state-space discretization (clustering) due to the inherent information loss. Also, Zeng and Smyth~\cite{DBLP:journals/neco/ZengGS93} noticed problems with the clustering approach. In this paper, we return to this open question and thoroughly study the validity of the clustering hypothesis. 
The results of our empirical study shall increase the confidence in existing and future work on the validation and verification of RNNs.

\begin{figure*}[t]
    \centering
    \includegraphics[width=0.7\textwidth]{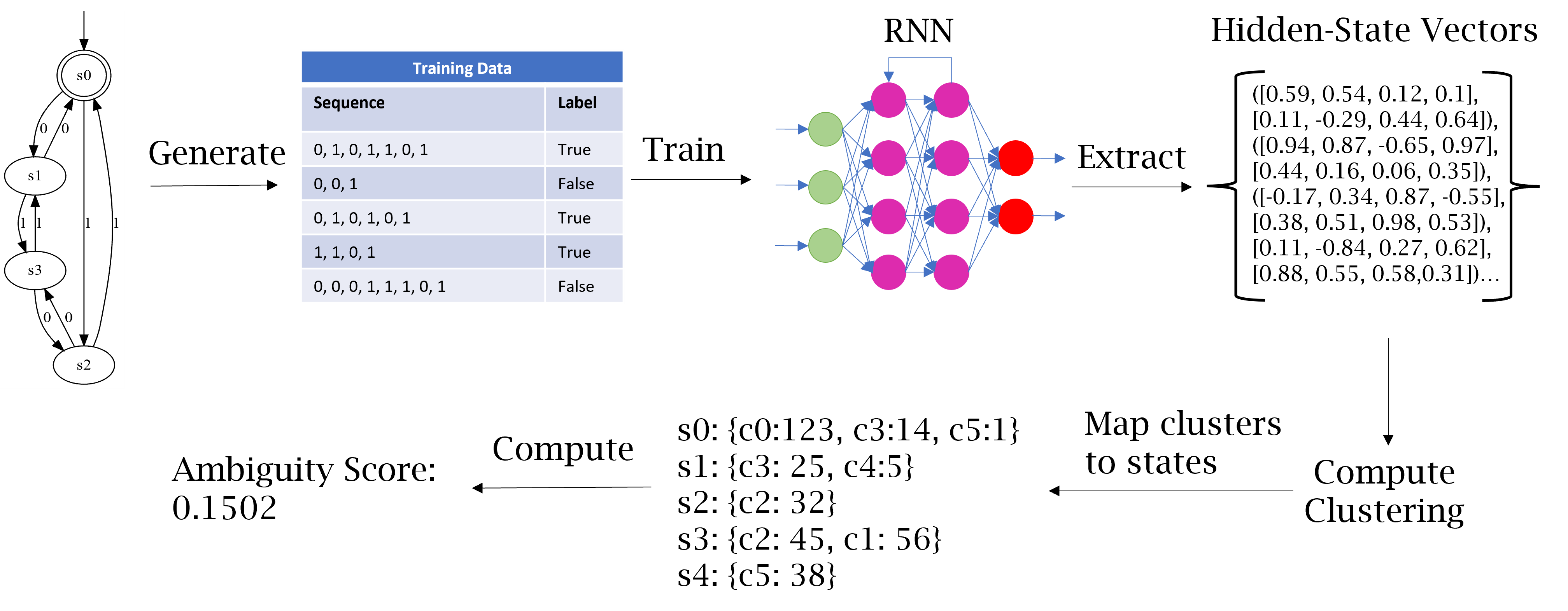}
    \vspace{-0.4cm}
    \caption{Overview of the process for assessing the quality of clustering functions.}
    \vspace{-0.5cm}
\label{fig:training_and_extraction}
\end{figure*}

% The aim of this paper is to revisit and examine the observation which states that an RNN's (hidden) state vectors form clusters. 
% We will refer to this hypothesis as the ``\textit{clustering hypothesis}''.

%The aim of this paper is to revisit and examine the clustering hypothesis in various settings, e.g., 
%considering different modern RNN architectures.
%In Sect.~\ref{sec:related_work}

% Old text: continue changes here (BKA)
%Below we outline several notable approaches that had either observed or assumed the clustering hypothesis.

%Relating to their observations, we formulate research questions regarding the clustering hypothesis.
More precisely, we investigate the following research questions: 
Are an RNN's hidden-state vectors linearly separable (RQ1)? 
Is the clustering hypothesis assumed for RNNs valid (RQ2)?
Do different training conditions affect clustering (RQ3)?
We approach these questions by training RNNs on regular languages with known minimal finite-state automata. By simulating the validation data on such RNNs we extract visited hidden states vectors, which are then used to compute multiple clustering functions. We analyze the correlation between derived clusters and automata states and assign an ambiguity score to each clustering configuration. Ideally, each cluster (or a set of clusters) should correspond to a unique state in the respective ground-truth automaton. An overview of this process can be seen at Fig.~\ref{fig:training_and_extraction}

%We analyze if this assumption holds and how different training conditions affect this property. 
%In order to better
%understand the structure of the internal (hidden)
%state space of RNNs, we also examine the (piecewise linear)
%separability of the state space into regions corresponding to automata states.
%We hope that our findings could be used by 
%researchers in the future to better understand the 
%internal decision-making process of RNNs. 

The contributions of this work can be summarized as follows:
% (1) overview of visualization techniques for RNNs internal state space, 
(1) analyses of $1350$ trained RNNs w.r.t.\ separability of their state space,
(2) computation and analysis % and visualization 
of clusters over the hidden-state vectors, 
(3) application of all methods on %correct-by-construction 
close-to-perfect RNNs, 
%(4) examination of the training process of mutated correct-by-construction networks, 
(4) and a framework containing all presented functionalities developed with \textsc{AALpy}~\cite{aalpy}, PyTorch~\cite{NEURIPS2019_9015}, and scikit-learn~\cite{scikit}. To the best of our knowledge, this constitutes the first detailed empirical evaluation of the 
"clustering hypothesis".

% The paper is structured as follows. In the remainder of this section we will outline notable related work and point out how their work relates to and builds on the clustering hypothesis. In Sect.~\ref{sec:prelim} we present preliminaries required for a self-contained paper. Sect.~\ref{sec:analysis}
% contains techniques and accuracy metrics that we use
% in Sect.~\ref{sec:experiments} for an empirical evaluation of the ``clustering hypothesis''. %and RNNs ability to model regular languages. 
% In Sect.~\ref{sec:conclusion} we state conclusions and discuss future work.
 %The paper is structured as follows. 
 \noindent
 {\it Structure.} %In the remainder of this section we  outline work related  %and point out how it relates to the clustering hypothesis. 
 In Sect.~\ref{sec:prelim}, we give preliminaries. Sect.~\ref{sec:analysis} discusses our research
 questions in detail and presents techniques and accuracy metrics that we use
in Sect.~\ref{sec:experiments} for an empirical evaluation of the ``clustering hypothesis''. %and RNNs ability to model regular languages. 
In Sect.~\ref{sec:related_work} we discuss related work, and in Sect.~\ref{sec:conclusion} we draw conclusions and discuss future work.

\noindent
\textbf{Related Work.}\label{sec:related_work}
% In this section, we will outline the papers that, in some form or another, assume the clustering of the hidden-state space. We will focus on the parts of the presented papers that are relevant to our inquiry, and we recommend to an interested reader to further study various interesting methods presented by the papers.
Omlin and Giles~\cite{DBLP:journals/nn/OmlinG96} were among the first to 
% apply principles of 
% grammatical inference 
to mine rules from RNNs. %More precisely, 
They proposed an algorithm for DFA extraction from a second-order RNN trained to recognize a regular language by applying a clustering algorithm over the RNN's hidden-state space. 
They also visually analyzed the clustering of the RNN
states, postulating that they are well separated.
Similar works can be found in~\cite{DBLP:journals/neco/CleeremansSM89,DBLP:conf/nips/GilesMCSCL91,DBLP:conf/nips/WatrousK91}.
%\todo{references taken Kolen93, only skimmed the papers, but they seem fine}
Since then, RNN research made significant 
advancements and the clustering hypothesis was criticized~\cite{DBLP:conf/nips/Kolen93}. 
%This prompts to revisit 
%and analyze 
%the hypothesis on modern RNNs.
%(or as they call it ``a neuron space''). 
% Their algorithm works on the assumption that the hidden-state space of the RNNs forms clusters, and that they correspond to the states of the automaton used to train the RNN. Furthermore, they postulate that these clusters are ``well separated'' between states of the automaton. 
%\textit{We will examine their assumption on both modern RNN architectures, as well as on larger networks compared to their seminal work.}

Zeng and Smyth~\cite{DBLP:journals/neco/ZengGS93} observed that the ``clustering hypothesis''~\cite{DBLP:journals/nn/OmlinG96} becomes unstable as longer sequences are used to extract hidden-state vectors. They propose changing the training as a solution.
% , or as they put it ``\ldots the network forgets where the individual states are \ldots''. This is a consequence of the vanishing gradient problem, an inherent challenge of RNN training. 
%They proposed an approach to mitigate this limitation by training a network in such a way that it is stable with longer sequences. They achieve this by discretizing the hidden-state space, so that the internal representation of a state is encoded with isolated points in the hidden-state space. 
%\textit{We will examine the observations that clusters tend to intertwine for longer sequences}.   
Schellhammer et al.~\cite{DBLP:conf/conll/SchellhammerDTB98} trained an Elman RNN on a natural language processing (NLP) task and constructed a state-transaction diagram representing a grammar of the data set with the help of clustering. However, they considered RNNs with only two hidden neurons.
% They have performed graphical cluster analysis of the network consisting of 2 hidden neurons, thus avoiding the prepossessing (with a  dimensionality reduction technique) of hidden-state vectors as a prerequisite to visualization. They observed that activation tends to be clustered according to inputs, while data with high-frequency in the training set is dispersed in several sub-clusters. They, as well as previous approaches, managed to extract rules (encoded as a DFA) from observed clusters. It is important to note that this, as well as previous approaches, extracted a non-minimal DFA. %\textit{This could indicate that a single automaton state could be broken into several disjoint clusters. We will examine this hypothesis in our analysis.}
%
%An empirical evaluation of rule extraction performed by
Wang et al.~\cite{DBLP:journals/neco/WangZOXLG18} empirically evaluated %took a closer look
various conditions that might influence DFA extraction from second-order RNNs.
%They empirically examined previously discussed approaches and concluded that DFAs can be ``stably'' extracted from RNN state space even when trained with short sequences. 
Interestingly, they observed that rules extracted from RNNs were more precise than RNNs themselves in classifying longer sequences. 
% While that might indicate that the extracted DFA is more precise than the RNN from which a DFA was extracted, we postulate that this is simply a consequence of a ``approximately correct'' DFA/rules extraction.
% That is, extracted rule set/DFA encodes input-output behavior of the RNN that conforms to the grammar used for the data set generation, while not encoding the sequences that RNN classifies. We made similar observations in~\cite{DBLP:conf/ifm/MuskardinAPT22}. \textit{More detailed discussion and examination of this assumption will be done in the future work.} 
Dong et al.~\cite{DBLP:conf/kbse/DongWSZWDDW20} combined principles from passive stochastic automata learning, with abstraction achieved by clustering the hidden-state space of an RNN. 
%More precisely, they applied the k-means clustering algorithm over the hidden states observed while processing sequences from the training data set. This method mapped high-dimensional hidden-state vectors to a finite number of discrete clusters.
Clustering enabled adversarial data detection via probabilistic verification.
% \textit{Method used in this paper prompted us to further examine the ``clustering'' hypothesis.} 
Hou and Zhou~\cite{DBLP:journals/tnn/HouZ20} extracted automata from different types of gated RNNs with the help of clustering. 
% They trained gated RNNs, on regular languages and a text classification task, and proposed a method similar to previously discussed ones. That is, they extract an automaton based on state transitions of the clustered hidden state space. They observed that the states indeed do cluster, and based on those clusters they construct a DFA representing the language that was used to train the RNN. 
They also reason about the gating mechanism and its influence on the
clustering. However, their evaluation is restricted to two regular 
languages and a coarse abstraction for an NLP task, where they limit 
the number of clusters to just two. We improve upon their analysis
by considering a significantly larger number of study subjects, RNN types,
clustering approaches, and parameterizations, thus providing more 
 and nuanced insights into the clustering hypothesis. 
%In addition, they perform visualization of a high-dimensional hidden-state space similarly as performed in Sect.~\ref{sec:analysis}. \textit{In line with their work, we will visualize and compute clusters of gated RNNs (more specifically LSTMs and GRUs) and examine how well they cluster under different conditions.}
Michalenko et al.~\cite{DBLP:conf/iclr/MichalenkoSVBCP19} examined the relationship between hidden (Elman) RNN states and the states of DFAs used for training data generation. They learned decoding functions from hidden states to (sets of) DFA states and found that such (linear) functions exist, though some DFA states may need to be grouped. Their results suggest that ``supervised clustering'' may not enable perfect reconstruction of 
FSM rules, but rather non-deterministic approximations. In this paper,
we examined \emph{if unsupervised clustering enables 
reconstruction of the finite-state semantics of the concept that is learned}, thus taking a view that is closer to practice. 
% and that linear functions appear to be sufficient.
% While the existence of such functions is a positive result, the necessary
% grouping of automaton states points at a weakness of the clustering hypothesis,
% as it means that semantically different states cannot be distinguish in the RNN
% state space. We make use of the observation that adding nonlinearity does not improve
% the accuracy of these functions, thus we perform linear discriminant analysis to which we compare clusters.
%\textit{In this paper, we move one step further by not only computing such discriminant functions, but we also examine the mismatch between them and hidden state clusters derived without information about the ground truth automaton.} 

\section{Preliminaries}\label{sec:prelim}
% \todo{NICE help: https://arxiv.org/pdf/1810.10708.pdf}
\begin{definition}
A \textbf{\gls*{DFA} over alphabet $\Sigma$ 
is a tuple $A = (Q, q_0, \delta, F)$}, where $Q$ is 
a finite set of states, $q_0 \in Q$ is the initial state, $\delta : 
Q \times \Sigma \rightarrow Q$ is the transition function, 
$F \subseteq Q$ are the final states.
\end{definition}
\noindent
% \textbf{Deterministic Finite Automata.}
A \textbf{\acrfull*{DFA}} defines a regular language comprising its accepted words.
% the words that it accepts.
We extend the transition function as usual to arbitrary-length sequences
$s \in \Sigma^*$, i.e., $\delta(q,\epsilon) = q$ and $\delta(q,e \cdot s')
= \delta(\delta(q,e),s')$, where $\epsilon$ is the empty sequence, $e\in \Sigma$, $s' \in \Sigma^*$. A word $w \in \Sigma^*$ is accepted iff 
$\delta(q_0,w) \in F$. 

Moore machines extend \gls*{DFA} by producing an output from a discrete
output alphabet $O$ in every state defined by a function 
$\lambda : Q \rightarrow O$. We can view both \glspl*{DFA} and Moore 
machines as functions $m : \Sigma^* \rightarrow O$ with 
$m(w) = \lambda(\delta(q_0,w))$, where $\lambda(q) = q \in F$ for
\glspl*{DFA}. A function $m$ defines a classification problem on words, 
which has two classes in the case of $\glspl*{DFA}$. 
Note that a Moore machine defines a regular
language over the alphabet $\Sigma \times O$. We train \glspl*{RNN} on 
such languages.

\noindent
% \textbf{Recurrent Neural Networks.}
% \glspl*{RNN}
\textbf{\Acrfullpl*{RNN}} model sequential processes and are trained on sequential 
data. By manipulating an internal state, 
which we call the hidden state in this paper, they capture temporal dependencies across several time steps. 
Based on the current 
hidden state, outputs are computed, which may be discrete in classification
tasks or continuous in regression tasks.
This makes them well-suited for applications, such as, natural language processing~\cite{DBLP:conf/emnlp/0001DL16}, intrusion detection~\cite{DBLP:conf/fdse/BontempsCML16},
and time series forecasting in various domains. In this paper, we examine
Elman RNNs~\cite{elman},
LSTMs~\cite{DBLP:journals/neco/HochreiterS97}, and GRUs~\cite{gru}.
Below we provide the definition of one-layer Elman RNNs following the
PyTorch implementation~\cite{NEURIPS2019_9015}, while definitions of other RNN types can be found in the appendix.
\begin{align*}
% \intertext{Elman RNNs:}
h_t &= g(x_t W_{ih}^T + b_{ih} + h_{t-1} W_{hh}^T + b_{hh})  \text{ where } g \in \{\tanh, \mathrm{ReLU}\} \\
y_t &= f(h_t W_{ho}^T) 
\end{align*}

Elman RNNs manipulate the hidden state
$h_t$ at time step $t$ through a simple recurrent structure based on the 
previous hidden state $h_{t-1}$. The most common activation functions
for the recurrent layer(s) are $tanh$ and \emph{rectified linear units} (ReLUs).
The current output $y_t$ is 
computed through another neural network cell with an activation function 
$f$, which works analogously for LSTMs and GRUs. For classification,  
the softmax function is often used as activation. LSTMs and GRUs improve 
upon Elman RNNs that suffer from the vanishing/exploding gradient
problem, which makes long-term dependencies hard to 
infer~\cite{rnn_review}.

% \todo{Would remove or replace next paragraph with one sentence}
% Elman RNNs suffer from the vanishing/exploding gradient problem, which 
% makes their training difficult, so that long-term dependencies are hard to 
% infer~\cite{rnn_review}. LSTMs tackle this problem by introducing gates in their
% structure and a cell memory that can track dependencies over longer
% time intervals. The gates basically control, which parts of the data
% are stored, kept or forgotten. 
% GRUs improve upon LSTMs by using a simpler structure, which does not 
% use a dedicated cell memory and is more efficient to train, while also
% achieving high accuracy.

% We cannot go into detail on \glspl*{RNN}, but
% we will comment on specific aspects where relevant. 
We view an RNN as a pair of functions $(r,o)$, where
$r : \mathbb{R}^h \times \mathbb{R}^m \rightarrow \mathbb{R}^
h$ updates the hidden state based on the previous hidden state
$h_{t-1}$ and the current input $i_t$. We use a one-hot encoding
for discrete inputs from an alphabet of size $m$.
For simplicity, we also use this view for LSTMs by 
analyzing the state space spanned by the concatenation of the 
hidden state $h_t$ and the cell state $c_t$ of LSTMs. 
The output function $o : \mathbb{R}^h \rightarrow C$
maps the current hidden state to an output class in $C$. 
We train RNNs on sequences sampled from regular languages defined
by \glspl*{DFA} and Moore machines. In the former case, we have 
$C = \{\mathit{true},\mathit{false}\}$, where $\mathit{true}$ denotes
acceptance of the sequence processed so far.
In the latter case, we have $C = O$, where $O$ is an output alphabet 
and $o$ maps to the last output produced by the corresponding Moore
machine.

\noindent
\textbf{Multiclass Classification.}
We examine the correspondence between hidden state vectors  
and ground truth models represented by deterministic finite
automata. Given a finite sample of vectors in $\mathbb{R}^h$
labeled with corresponding automaton states from a set $Q$, we 
formulate a multiclass classification problem. We 
apply (generalized) linear models to solve the problem so 
that their accuracy provides insights about the (piecewise linear) 
separability of $\mathbb{R}^h$ into decision regions corresponding to
$Q$. 

In this setting, we have a sample from 
$\mathcal{S} \subset \mathbb{R}^h \times Q$. A pair 
$(h,q) \in \mathcal{S}$, contains a vector $h$ in
the hidden state space of an RNN and an automaton state $q$, which
is the class label for the classification problem. Our goal
is to find a function $cl : \mathbb{R}^h \rightarrow Q$ that 
classifies vectors correctly. We apply two representative 
techniques for classification: \gls*{LDA} and \gls*{LR}.
Both techniques yield linear models, meaning that their output classes
depend linearly on the hidden state vectors serving as inputs.
\gls*{LR} is actually a generalized linear model since it uses a
nonlinear transformation of the input space. However, for more 
information we refer to the
literature~\cite{DBLP:books/lib/Bishop07}.

\noindent
\textbf{Clustering.}
Clustering deals with the identification of structure in data, an important problem in unsupervised learning~\cite{clustering_overview}. In contrast to
multiclass classification, clustering techniques take 
unlabeled data points as inputs. Clustering
groups data points into clusters, such
that ideally data points in the same clusters are \emph{similar}
to each other and data points from different clusters
shall be dissimilar from each other. The notion of
similarity is generally a problem-specific measure.
The input to a clustering technique is an unlabeled sample 
of data points $\mathcal{US}$, in our case: $\mathcal{US} \subset \mathbb{R}^h$. 
Thus, we can view a clustering as a function $c : \mathcal{US} 
\rightarrow K$ assigning cluster labels to data points.
% For instance, the Euclidean distance may be used in 
% $\mathbb{R}^h$.
%A vast amount of clustering techniques can be found in the literature with different characteristics. 
In this paper, we focus on popular, efficient techniques 
available in the library scikit-learn~\cite{scikit}.
We apply k-means~\cite{kmeans}, a partitional clustering 
technique~\cite{clustering_overview} and three density-based
clustering techniques: mean shift~\cite{meanshift}, DBSCAN~\cite{dbscan},
and OPTICS~\cite{optics}.

\textbf{K-means} partitions data 
into $k$ clusters, assigning each data point $p$ to
the cluster whose center is closest
to $p$. The center is the mean of all data points in a cluster. 
K-means creates piecewise linear decision boundaries, where the distance from 
centers of a neighboring cluster is equal. 
Density-based techniques creates clusters of arbitrary shape that
are identified as regions with a 
high density of data points. 
\textbf{DBSCAN} defines regions as clusters if at least $\mathit{minNeighbors}$
points are in a neighborhood of a size defined by a radius $\epsilon$. 
By merging clusters that 
are close to each other, it finds arbitrarily-shaped clusters.
\textbf{OPTICS} follows the same basic approach as DBSCAN, but improves
it by mitigating the sensitivity on the neighborhood size
$\epsilon$~\cite{clustering_survey}. Like k-means, \textbf{mean shift}
clusters a dataset based on centroids. Conversely, however, it attempts
to find maxima in the density function underlying the distribution 
of the data and does not require a preset number of clusters $k$. It uses a bandwidth parameter $bw$ to define regions for
the update of centroids.
As k-means creates piecewise linear decision boundaries, 
it is the closest among the considered
clustering techniques to \gls*{LDA}.

% However, \gls*{LDA} does not only 
% consider distances between class centers. It also maximizes 
% the ratio of the covariance across different classes and the covariance within
% individual classes, when data is projected to decision boundaries.

% \subsubsection{Verifying RNN Accuracy}

% To verify the RNN generalization capabilities, we will perform the approach outline in~\cite{DBLP:conf/ifm/MuskardinAPT22}.
\section{Method}\label{sec:analysis}
This section presents the basis for the analysis of the ``clustering hypothesis''. 
%Firstly, we introduce research questions setting that we consider and the setting for the analysis.
% the regular languages that were used to generated data to train RNNs and outlay the RNN (hyper)parameters and the training setup. 
Firstly, we formulate our research questions and introduce the setting.
Then, we provide details on the analysis %performed in the analysis 
including accuracy measures. % that we employ. 

\subsection{Research Questions}
\noindent
\emph{Setting.}
% We begin by introducing the basic setting for RNN training and data collection.
In our experiments, we train RNNs on regular languages over 
an alphabet $\Sigma$.
For each regular language $L \subseteq \Sigma^*$, we sample a dataset
$D \subset \Sigma^*$.
% For each prefix $p$ of every word
% in $D$ we also determine its label $L(p) = p \in L$, which denotes whether
% $p$ is in the language under consideration.
We split the sample $D$ into training data $D_t$ 
%containing $n_t$ words 
and validation data $D_v$ of sizes $n_t$ and $n_v$. % containing $n_v$. 
The former is solely used for training
whereas the latter provides a stopping criterion for training and data for the
analysis of the clustering hypothesis.

For the remainder of this section, we fix a regular language $L \subseteq \Sigma^*$
and a sample dataset $D  \subset \Sigma^*$. Let $A = \langle Q,q_0,\delta,F \rangle$
be a minimal DFA accepting exactly $L$ and 
% \paragraph{Data Collection.}
let $R=(r,o)$ be an RNN over the hidden state space $H = \mathbb{R}^h$ trained to 
recognize $L$. 
By processing every word in $D_v$ simultaneously with the recurrent part $r$ of $R$ and 
the minimal DFA $A$, we determine the hidden states traversed by $R$ as well as the 
corresponding automaton states reached by $A$. We store these data as
pairs $(h,q) \in H \times Q$. For the remainder of this section 
let $\mathcal{HQ} \subset H \times Q$ be a concrete sample of such pairs and 
let $\mathcal{H}$ be the same sample without states $Q$. 
% High-level overview of an algorithm mapping hidden states to automaton states can be found in Alg.~\ref{alg:hidden_state_collection}.
Algorithm~\ref{alg:hidden_state_collection_2} formalizes the creation of $\mathcal{HQ}$.

% \begin{algorithm}[t]
% \caption{Mapping of automaton states to hidden states}
% \label{alg:hidden_state_collection}
% \begin{algorithmic}[1]
% \Require ground truth automaton $aut$, RNN trained to recognize ground-truth automaton $rnn$, validation data-set $D_v$
% \Ensure map from automaton states to hidden-state vectors $\mathit{stateHiddenVectorMap}$
% \State $\mathit{stateHiddenVectorMap} \gets \mathit{map\{hiddenStateVector : automatonState\}}$
% \ForAll {$(sequence, label)~\mathbf{in}~D_v$}
% \State $\mathit{aut.resetToInitialState()}$
% \State $\mathit{out, hs} \gets \mathit{rnn.reset()}$
% \ForAll {$i~\mathbf{in}~sequence$}
% \State $\mathit{aut.step(i)}$
% \State $\mathit{out, hs} \gets \mathit{rnn.step(i, hs)}$
% \State $\mathit{q}~\gets~aut.currentState$
% \State $\mathit{stateHiddenVectorMap[hs]} \gets stateHiddenVectorMap[hs] \cup \{q\}$
% \EndFor
% \EndFor
% \State \Return $\mathit{stateHiddenVectorMap}$
% \end{algorithmic}
% \end{algorithm}

\begin{algorithm}[t]
\caption{Labeling of hidden states with automaton states.}
\label{alg:hidden_state_collection_2}
\scriptsize
\begin{multicols}{2}
\begin{algorithmic}[1]
\Require $A = \langle Q,q_0,\delta,F \rangle$, RNN $(r,o)$, initial hidden state $h_0$, validation data $D_v \subset D$
\Ensure pairs of hidden and aut.\ states $\mathcal{HQ}$
\State $\mathcal{HQ} \gets \emptyset$
\ForAll {$(w, label)~\mathbf{in}~D_v$}
\State $q \gets q_0$, $\mathit{hs} \gets h_0$
\State $\mathcal{HQ} \gets \mathcal{HQ} \cup \{(hs, q)\}$
\ForAll {$i~\mathbf{in}~w$}
\State $q \gets \delta(q,i)$
\State $\mathit{hs} \gets \mathit{r(hs,\rho(i))}$ \Comment{$\rho$ encodes inputs}
\State $\mathcal{HQ} \gets \mathcal{HQ} \cup \{(hs, q)\}$
\EndFor
\EndFor
\State \Return $\mathcal{HQ}$
\end{algorithmic}
\end{multicols}
\end{algorithm}

% add to next section
% Then we continue to analyze the hidden-state space with known clusters, computed clusters. We also extend our analysis to the computed correct-by-construction RNNs.

\noindent \textbf{RQ1: Are an RNN's hidden-state vectors linearly separable?}
As a first step, we investigate whether hidden-state vectors can be (piecewise linearly) separated
into regions corresponding to automaton states. 
We formulate this as a multiclass classification problem based on $\mathcal{HQ}$,
the sampled hidden-state vectors labeled by automaton states.
That is, we train a classification model on $\mathcal{HQ}$ to learn a
function $cl : \mathbb{R}^h \rightarrow Q$ and we evaluate its accuracy at classifying
hidden states correctly. 
% While investigating RQ1, we will observe differences between different clustering algorithms
% and across different regular languages serving as experimental subjects. Some regular
% languages may lead to automata structures that are harder to learn, e.g., parity languages (Tomita 5 in Sect.~\ref{sec:eval}) have been used to establish lower bounds on neural network size~\cite{DBLP:journals/neco/CarrascoFVN00}, such that the data available to clustering 
% algorithms is insufficient to compute meaningful clusters. 
%  For this purpose, we compute discriminant functions via \gls*{LDA} from $\mathcal{HQ}$, the same samples of the hidden state space that are also provided to the clustering algorithm, but labeled with corresponding automaton state. 
We rely on (generalized) linear models \gls*{LDA} and \gls*{LR} for this task. 
This choice is motivated by the observation that non-linear functions were not
required in a similar setting
%explored by Michalenko et al.
~\cite{DBLP:conf/iclr/MichalenkoSVBCP19}.
While \textit{piecewise linear separability} and low classification error of linear models 
\textit{do not imply that the hidden states form well-defined clusters}, we
gain insights into the structure of the hidden-state space. 
For example, if piecewise linear separation is not possible then k-means
will generally not find accurate clusters.
Hence, by using linear models for separability 
we actually enable a fairer comparison to
clustering. 
% K-means, one of the most popular clustering techniques, 
% uses piecewise linear decision boundaries, like \gls*{LDA}. 
% The computed discriminant functions provide an insight as to whether decision boundaries
% can be established that cleanly separate the automaton states, when using additional
% knowledge in the form of state label.  

\noindent
\textbf{RQ2: Is the clustering hypothesis assumed for RNNs valid?}
The main research question concerns the validity of the clustering hypothesis. 
To examine it experimentally, we consider two aspects 
related to the clustering of RNN states.
% we first need to discuss two related research questions.
\begin{compactitem}
    \item \textbf{The definition of a cluster.} %\textbf{How to define a cluster? } 
    Works based on the hypothesis
    apply different clustering techniques, 
    %potentially preceded by dimensionality reduction
    %of the hidden state space. 
    hence we will analyze the effect of
    different popular clustering techniques.
    \item \textbf{Usefulness of a cluster.} The appeal of clustering 
    stems from its potential as an abstraction for RNNs, thus enabling 
    model-based reasoning~\cite{DBLP:conf/kbse/DongWSZWDDW20}. Hence, clusters should
    possess similar qualities as abstractions for software systems. Concrete hidden states \emph{abstracted to the same clusters} should behave similarly. Moreover, 
    an abstraction should be small enough to enable efficient subsequent analyses, 
    thus we will examine the number of detected clusters.
\end{compactitem}
We will analyze these aspects based on experiments with regular languages, where we know the 
minimal ground-truth automaton. Hence, answering \textbf{RQ2} in this context can also be formulated as: \textbf{Do hidden state clusters correlate with states of an automaton accepting the same language?}

\noindent\textbf{RQ3: Do different training conditions affect clustering?}
% The training of RNNs is governed by a large number  of parameters and aspects, 
% which potentially affect clustering.
RNN training parameters and initialization potentially affect clustering, therefore we investigate
a different training initialization. We create RNNs from ground-truth 
automata as described below. These RNNs are guaranteed
to form dense clusters. After introducing 
noise into the RNN weights, we will examine
if training adjusts the weights in a way that 
again leads to well-formed, meaningful 
clusters.

% To keep the configuration manageable, we will examine two high-level changes 
% of the training, adapting the initialization of the network and the stopping criterion.

% \emph{Initialization.} 
% In Sect.~\label{sec:correct_by_construction} we describe the computation of  correct-by-construction RNNs. Such RNNs are created so that they conform exactly to the automaton. Their hidden-state space is saturated and forced in clusters that correspond to
% states of the automaton. We will examine the following: After introducing Gaussian noise in the correct-by-construction RNNs weights and retraining them, does their hidden-state space still gravitate toward the initially specified, ``perfect'' conditions. That is, we will examine if training causes clusters to form again after the introduction noise changed the hidden state space.

\subsection{RNN Construction}\label{sec:rnn_construction}
Our construction of RNNs to encode automata
is similar to those proposed in \cite{DBLP:journals/neco/AlquezarS95,DBLP:journals/tnn/GoudreauGCC94,minsky_book}, therefore we keep the presentation brief and 
provide the complete construction in the appendix. 
We use Elman RNNs with $\mathit{tanh}$ activation and a single layer, 
which we use in the saturated area of 
$\mathit{tanh}$, where it works like a threshold gate.
We implement $\delta$ of the ground-truth DFA 
by setting $W_{ih}$, $W_{hh}$, $b_{ih}$, and $b_{hh}$, and
we implement $q \in F$ by setting output-layer weights $W_{oh}$.

\begin{figure}[t]

\centering
\includegraphics[width=.33\textwidth]{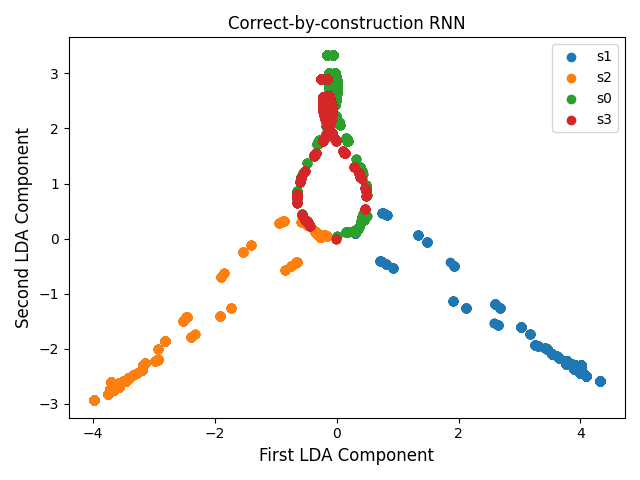}\hfill
\includegraphics[width=.33\textwidth]{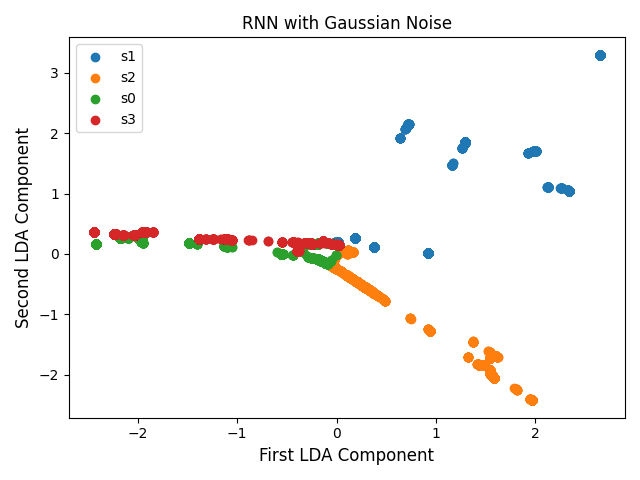}\hfill
\includegraphics[width=.33\textwidth]{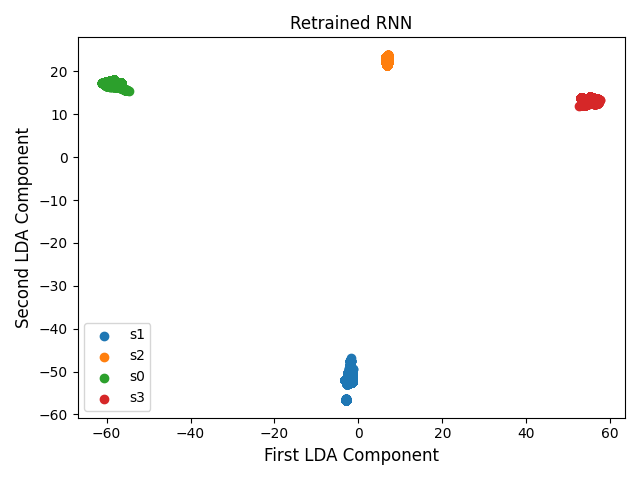}

\vspace{-0.2cm}
\caption{Visualization of the hidden-state vectors of a correct-by-construction RNN(left), constructed RNN with Gaussian noise (middle), and trained noisy network (right), where we apply LDA for dimensionality reduction to 2D.  %X and Y axis are the first and second components obtained with LDA's dimensionality reduction.
    }
\vspace{-0.5cm}
\label{fig:retraining_of_constructed_RNN}

\end{figure}

An RNN encoding an automaton has a state space with dense clusters
corresponding to transitions, i.e., such an RNN would form
$|Q| \cdot |\Sigma|$ many clusters. Hence, we need to 
analyze whether RNNs learn such non-minimal 
representations. Attempting to further train
such an RNN would usually stop immediately,
therefore we add noise to the 
weights $W_{ih}$, $W_{hh}$, $b_{ih}$, and $b_{hh}$
to change the state transitions. This lets us analyze the effect
of training \emph{close-to-perfect} RNNs. We use Gaussian noise
with a mean of $0$ and a standard deviation of $wn$. Figure~\ref{fig:retraining_of_constructed_RNN} shows a visualization of a constructed RNN, before and after introducing noise, and after training the noisy RNN. Here, we apply LDA for dimensionality reduction by projecting 
the data to a 2D-subspace that provides the best separation of data according 
to the \gls*{LDA} criterion.
%where we map the states to the two \gls*{LDA} components
% that provide the best separation. 
%In the left figure (without noise), some states can be clearly separated, though not all due to dimensionality reduction, while there is a less clear separation in the middle figure (with noise). The figure on the right shows that clusters are formed after training the noisy RNN.  

\subsection{Accuracy Metrics}\label{sec:accuracy_evaluation} %\todo{Would word this subsection differently. Maybe something like Accuracy Metrics or Defining Accuracy for RNN and Clustering}
We discuss the evaluation of the accuracy of RNNs and clusterings below. 
The former is based on the misclassification
of words w.r.t.\ the ground-truth automaton $A$. 
Clustering accuracy is based on the ambiguity 
resulting from interpreting clusters as states of a finite-state model, compared to the states of $A$. 

\noindent
\textbf{RNN Accuracy Validation.}
We validate the accuracy of an RNN $R = (r,o)$ by sampling words
$\Sigma^*$ to form an accuracy-validation set $AV$ and 
checking whether $R$ agrees with the ground-truth $A$, i.e., accepts
the language $L$.
% This is an approximate check if $R$ accepts the language $L$. 
For each word $w \in AV$ we process $w$ with $R$ by 
repeatedly applying $r$
to get the final hidden state $h_{|w|}$ and computing the RNN output via $ro = o(h_{|w|})$. Then, we check whether $ro$ agrees with the ground truth, 
where we define $\mathit{agree}_{R,L}(w) =_\mathit{def} ro \leftrightarrow w \in L$. 
We define the accuracy of $R$ as $\mathit{Acc}_R(AV) = \frac{|\{\mathit{agree}_{R,L}(w)\mid w \in AV\}|}{|AV|}$. 

\noindent
\textbf{Measuring the quality of clustering.}\label{sec:ambiguity_computation}
We use the following intuition to define the optimality of clustering:
\emph{All data points in a cluster should correspond to a unique 
state in the ground truth $A$}. 
Formally, we can view a clustering as a function $c : \mathcal{H} 
\rightarrow K$ mapping sampled hidden states to cluster labels. This
lets us compare a clustering $c$
to the optimal mapping $hq : H \rightarrow Q$ based on the samples in 
$\mathcal{HQ}$.
To relate $c$ and $hq$, we define the clustering of $c$ 
as optimal if $hq = \alpha \circ c$ holds,  where $\alpha : K 
\rightarrow Q$ renames cluster labels to states.
Such a $c$ allows extracting a DFA with states $K$ from an RNN $R$
that is equivalent 
to ground truth $A$ if $\mathit{Acc}_R(AV)=1$ and $AV$ is large enough.
To empirically evaluate a concrete $c$, we define the mismatch 
between $c$ and $hq$ based on entropy. 

Before formally defining the mismatch between $c$ and $hq$, which we
term \emph{ambiguity} of $c$, let us analyze criteria for good versus 
bad clusters. To apply clustering as an 
 abstraction over $H$, a cluster should group semantically similar
 states. Conversely, semantically different states should not 
 be in the same cluster.
Hence, we can derive that a clustering with $|K| < |Q|$ cannot be 
optimal, since all states in a minimal DFA are different, 
i.e., distinguishable through future behavior.
In other words, a DFA extracted from the corresponding RNN 
with states identified by clusters $K$
would not be equivalent to the ground-truth DFA $A$.
If $|K| \geq |Q|$, the clustering may have a small mismatch, 
but the RNN may have learned a non-minimal representation
if $|K| > |Q|$. The latter would not be desired for abstraction due to 
efficiency reasons. 
Next, we consider the best- and worst-case distributions of states
in a cluster. Ideally, all hidden states in a cluster should map to the
same automaton state. The worst case is achieved by a 
uniform distribution over states, as then the cluster
would possess no semantic relation to the concept that has been learned.

We define ambiguity via the well-known concept of \emph{information
entropy}, which measures the degree of uncertainty (mismatch) in our
clustering mappings. The entropy of our best-case clustering would be
$0$, while the worst-case clustering (uniform distribution) would give 
maximal entropy. Given, the general definition of entropy
$
Entropy(X) =_{\textit def} - \sum_{x \in \mathcal{X}} p(x) \log p(x)   
$ 
with $X$ being a random variable with outcomes in alphabet $\mathcal{X}$
with distribution $p: \mathcal{X} \rightarrow [0,1]$,
we can instantiate it to measure the uncertainty in our cluster mappings via:

%Entropy treats these cases as two extremes, where the former is mapped to the value of zero,  
%thus we use it as a basis for ambiguity of a cluster:
\begin{align*}
    \mathit{amb}(k) &= - \sum_{q \in Q} 
    \frac{n_{q,k}}{n_k} \log_{|Q|} \frac{n_{q,k}}{n_k} \text{ where} \\
    n_{q,k} &= |\{(h,q) \mid (h,q) \in \mathcal{HQ}, c(h) = k\}| \text{ and } 
    n_{k} = \sum_{q \in Q} n_{q,k} \\
    \mathit{amb}(c) &= \frac{\sum_{k \in K}\mathit{amb}(k)}{|K|} 
    \qquad\qquad
    \mathit{wamb}(c) = \frac{\sum_{k \in K}\mathit{amb}(k) \cdot n_k}{|\mathcal{HQ}|} 
\end{align*}

By using the logarithm with base $|Q|$ we ensure that the ambiguity is normalized to the interval
$[0,1]$. For the ambiguity of a clustering function, we compute the average of 
all clusters and the weighted average $\mathit{wamb}$. 
We noted above that ideally $hq = \alpha \circ c$ for a renaming $\alpha$. This is achieved iff $\mathit{amb}(c) = 0$. We say that clustering is perfect if 
it achieves a (weighted) ambiguity of zero. 

% \begin{proposition}
% Given an RNN $R = (r,o)$ over hidden state $\mathbb{R}^h$, sampled hidden states $\mathcal{H}$
% containing initial hidden state $h_0$, 
% a clustering $c :  \mathcal{H} \rightarrow K$, a ground-truth automaton $A$, 
% and a large enough accuracy-validition set $AV$: If $\mathit{Acc}_R(AV) = 1$
% and $\wamb(c) = 0$ then we extract a DFA $RA=(K, q_{0R},\delta_R, F_R)$ from $\mathcal{H}$ and $c$ that is 
% equivalent to $A$.
% \end{proposition}

% \begin{proof}
% For the construction, set $q_{0R} = c(h_0)$, $\delta(c(h_{t-1}),e) = c(h_{t}))$
% for two hidden state $h_{t-1}$ and $h_t$ sampled consecutively when processing $e$ with $R$,
% and $k \in F_R$ if all $o(h) = \mathit{true}$ for all $h\in \mathcal{H}$ with $c(h) = k$.
% Suppose for now that $AV = \Sigma^*$. Then we have for all 
% $k \in K$ either $o(h) = \mathit{true}$ for all $h$ such that $c(h) = k$ or $o(h) = \mathit{true}$
% for all such $h$. Since $\mathit{Acc}_R(AV) = 1$, the RNN needs to agree with $A$
% on all words. 

% Now we can shows by induction the above proposition by induction. 
% For the base consider $h_0$ and $q_0$
% \end{proof}

To enable a straightforward comparison between classification 
models and unsupervised clustering, we apply ambiguity also for 
classification models, by interpreting predicted classes as cluster
labels. Note that ambiguity of zero coincides with a  misclassification rate of zero. Thus, zero ambiguity of linear models implies (piecewise linear) separability on the sample dataset $\mathcal{H}$.

Alternatively to our notion of ambiguity, 
we could also use normalized mutual information (NMI), 
a commonly used estimate of clustering quality with an information-
theoretic interpretation. However, clustering size affects NMI such that
a perfect, but slightly non-minimal clustering
may have a lower NMI than an ambiguous, but 
small clustering. Since the former (a perfect, non-minimal clusterings)
is likely more useful for RNN analyses than the latter (imperfect clusterings), we focus on ambiguity and examine clustering size separately.

% In the case of $|K| = |Q|$ such an $\alpha$ should be a bijection. Hence, a stricter
% ambiguity $\mathit{amb}_\alpha(c)$ would be possible by defining 
% $n_{q,k}$ as $|\{(h,q) \mid (h,q) \in \mathcal{HQ}, c(h) = k, \alpha(k) = q\}|$, where $\alpha$ is bijective, is the same for all clusters, and minimizes $\mathit{amb}_\alpha(c)$.
% However, this definition is less efficient to compute, does not 
% readily work for $|K| > |Q|$,
% and $\mathit{amb}(c) = 0$ iff $\min_\alpha \mathit{amb}_\alpha(c) = 0$.

% and their effect on the clustering accuracy. We configure the number of clusters $k$ for k-means depending on the number of states $n$ of the ground-truth automaton. . Finally, for mean shift we use different bandwidth values $\mathit{bw}$, which determines the size of found clusters. 

% Following the procedure applied in~\cite{DBLP:conf/kbse/DongWSZWDDW20}, we extract hidden-state vectors from the
\section{Evaluation}\label{sec:experiments}
In this section, we present the experimental setup and results on the clustering hypothesis to answer the research questions defined in Sect.~\ref{sec:analysis}. The code required to reproduce all experiments can be found at~\footnote{\url{https://github.com/DES-Lab/Clustering_RNN_hidden_state_space}}.

\subsection{Experimental Setup}
\noindent
\emph{Case Study Subjects.}
We performed experiments with 59 regular languages encoded by automata models. Five of the languages are evaluation subjects from the literature, including three of the Tomita grammars~\cite{tomita:cogsci82} (Tomita 3, 5, and 7), 
a Moore machine model of an MQTT server~\cite{DBLP:conf/icst/TapplerAB17}, %learned in previous work (Double blind violation) 
and a regular expression used by Michalenko et al.~\cite{DBLP:conf/iclr/MichalenkoSVBCP19}.
Additionally, we randomly generated 30 DFAs and 24 Moore machines with 
up to $12$ states and $72$ transitions using \textsc{AALpy}~\cite{aalpy}.
%The random DFAs have five or ten states, and alphabet sizes of two, four, or six.
%For each combination of state and alphabet size, we generated five DFAs. We generated three Moore machines for each combination of %eight and twelve states, two and four inputs, and three and five outputs.\todo{Commented out here.}
%Note that Moore machines encode regular languages that contain all possible input-output sequences. 
%From an RNN perspective, they enable us to analyze how multi-class classification tasks affects clustering. 

\noindent
\emph{Training.}
%let $R$\todo{introduce proper notation
%for RNNs} be an RNN with hidden state space $\mathbb{R}^n$. 
For each regular language, we trained an RNN to achieve perfect accuracy for 3 consecutive epochs on the validation data to potentially increase the likelihood of forming clusters in the hidden state space~\cite{DBLP:journals/corr/abs-2201-12451}.
We trained Elman RNNs~\cite{elman} with $tanh$ and $\mathit{ReLU}$, LSTMs~\cite{lstm}, and GRUs~\cite{gru}, each of them with one layer of size $n$, one layer of size $1.5 n$, two layers of size $n$, where $n$ is the number of transitions of the ground-truth automaton producing the language under consideration. We have chosen these network sizes, since a one-layer Elman RNN with $n$ hidden neurons is sufficient to encode automata with $n$ transitions, as in our RNN construction (see Sect.~\ref{sec:analysis}).
Additionally, we also train slightly larger networks, as an increased size may be beneficial for training. 
We decided to not consider very large networks, as it would complicate the clustering analysis due to dimensionality reduction becoming more important, hindering us from concentrating on clustering approaches.
For all experiments, we used the ADAM optimizer~\cite{DBLP:journals/corr/KingmaB14} with a learning rate of $0.0005$. The training data consisted of $n_t=50k$ randomly sampled words of lengths in the range $[1,15]$, with labels derived from the ground-truth model. The validation data contained $n_v=2000$ words, resulting in appr. $10k$ different hidden-state vectors for clustering. For all experiments, we trained two RNNs per configuration (network type, size). 

\noindent
\emph{Classification \& Clustering.}
We used \gls*{LDA} and \gls*{LR} to learn classification
models to determine whether automaton states can be %(piecewise linearly)
separated in the hidden state space. Both approaches are supervised
classification techniques, to which we provide $\mathcal{HS} \subset 
H \times Q$, sampled pairs of hidden states and automaton states. %a sample of the mapping from
% hidden-state vectors in $H$ to their corresponding automaton states $Q$ serving as labels. 
% An automaton state can be viewed as a label of a hidden-state vector in a multiclass classification problem. Classification are 
% usually evaluated in terms of misclassification rate. For comparability 
% with clustering, we compute the ambiguity
% measure \gls*{LDA} and \gls*{LR} by using the
% predicted class of a hidden state vector
% as its cluster label in $K$. This convention 
% ensures that the misclassification rate is 
% zero iff the ambiguity of a classification 
% function is equal to zero. Hence, an ambiguity
% of zero implies linear separability.
%

For the clustering algorithms, we apply 
various parameterizations and use
Euclidean distance as a distance metric. 
We set the $k$ of k-means based on
the size $n = |Q|$ of the 
minimal ground-truth automata $A$ with 
$k \in \{n-1, n, n+1,
2n, 4n, 6n, 8n\}$. With the first 
three values, we check whether a 
good clustering exists that is close to the
minimal automaton representation. We also use 
greater values because an RNN may 
learn a non-minimal representation, like in 
our construction. The evaluation languages
have up to $6$ symbols, thus 
$k=6n$ would be sufficient for our
construction of RNNs.
We choose the parameters of DBSCAN
and \emph{mean shift} as follows. For DBSCAN we experiment with multiples 
of the neighborhood size $\epsilon=0.5$, the default in the
scikit-learn library. We leave the other parameter 
$\mathit{minNeighbors}$ at its default value of $5$.
For mean shift, we estimate the
bandwidth $bw$ with scikit-learn, which we denote 
$\alpha$, to perform experiments with multiples of 
$\alpha$. As OPTICS improves upon
DBSCAN by mitigating its sensitivity on 
parameter values, we only apply its default 
parameterization.
Since mean shift and OPTICS require more computation time than other techniques, we reduced the sample $\mathcal{H}$ to $25\%$ of its size for these two.

% \subsection{Experiments Performed on All Network Types}\todo{Maybe better titles for sections, or we can have 1 section with 2 subsections}
\subsection{Overview of All Experiments}
In the first set of experiments, we consider all four RNN types, Elman RNNs with ReLU and tanh activation functions, LSTMs, and GRU networks, as well as all clustering techniques. We evaluated all RNNs and only considered those whose accuracy ($\mathit{agree}_{R,L}$ from Sect.~\ref{sec:accuracy_evaluation} evaluated on accuracy-validation data) was $\geq 80\%$. We chose this accuracy cutoff as, in practice, RNNs rarely achieve perfect generalization due to the complexity of the underlying task or quality of training data. In total, this resulted in a selection of $1350$ from $1416$ RNNs.

\begin{figure}[t]
    \centering
    \input{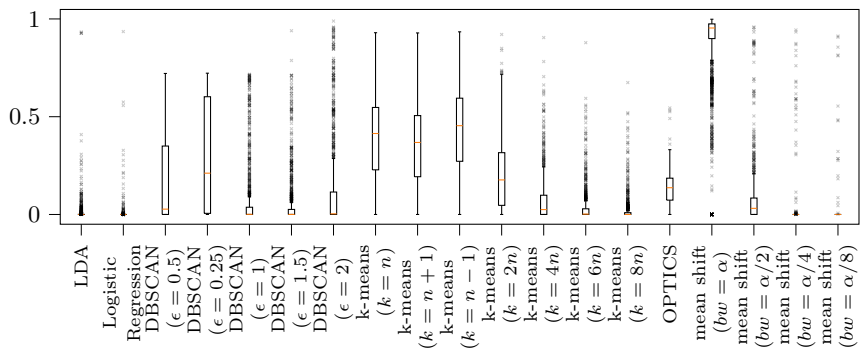}
\vspace{-0.3cm}
    \caption{Boxplots of the weighted ambiguity resulting from different 
    clustering techniques for all $1350$ experiments whose RNNs achieved at least 80$\%$ accuracy.}
\vspace{-0.3cm}
    \label{fig:clustering_all_80_percent_acc}
\end{figure}

Figure \ref{fig:clustering_all_80_percent_acc} summarizes the results of these experiments. It shows boxplots of the weighted cluster ambiguity $\mathit{wamb}$ resulting from different clustering techniques, with outliers denoted by small crosses. Furthermore, the second row of Table~\ref{tab:number_of_perfect_clustering} shows the number of perfect clusterings, i.e. $wamb = 0$, achieved by each method.

\begin{figure}[t]
    \centering
    \input{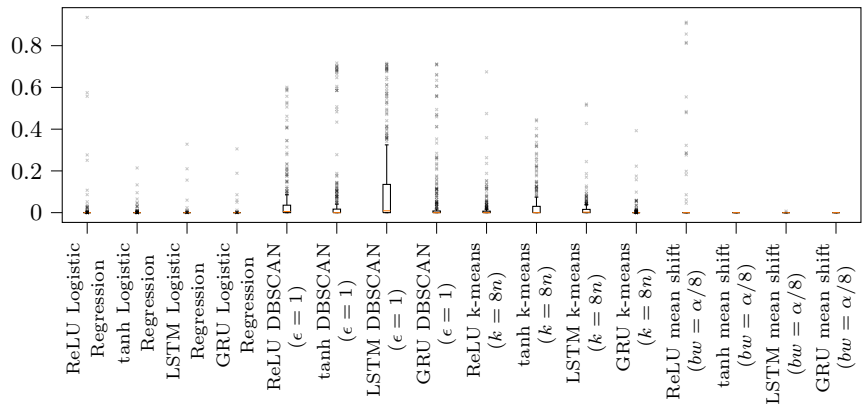}
    \vspace{-0.3cm}
    \caption{Weighted ambiguity for selected methods sorted by network types.}
    \vspace{-0.3cm}
    \label{fig:different_rnn_types}
\end{figure}

We observe that the LDA and \gls*{LR} are able to achieve perfect classification in 74\% and 91\% of the cases, respectively. LDA has mean $wamb$ of $0.0009 \pm 0.05$, while \gls*{LR}'s mean was $0.0004 \pm 0.039$. Relating to \textbf{RQ1}, this high level of accuracy indicates that in the majority of considered cases, even with networks that did not achieve perfect accuracy, \emph{piecewise linear separability of the hidden state space is possible}. Below we will show that the level of accuracy increases for perfectly trained RNNs. 
Recall, however, that LDA and \gls*{LR} are derived using additional information about the correspondence between hidden and automaton states, which is usually not available, therefore we examine the ambiguity of unsupervised clustering approaches.

%Let us continue with the analysis of k-means. Knowing that LDA and \gls*{LR} split the hidden state space in $n$ regions, initializing k-means with number of clusters $k$ set to $n$ would be sufficient to achieve similar ambiguity score. However, that was not the case. This indicates that k-means depends heavily on the initial cluster center positions. To mitigate this, we increased the number of clusters and observed that increasing $k$ strongly correlates with the accuracy of clustering. K-means with $k=8n$ was able to find perfect clustering in 58\% cases. With such perfect clustering, we can extract a non-minimal automaton of size $8n$, which can then be minimized to the ground-truth model. We will reason more about cluster sizes in Sec.~\ref{sec:cluster_sizes}

The median weighted ambiguity of DBSCAN with $\epsilon\geq1$ is almost equal to~$0$, which means that in at least half of the cases, we can identify semantically meaningful discrete states from clusters.
Likewise, k-means achieves a low ambiguity when given at least four times as many clusters as ``necessary'', i.e., $k\geq 4n$. This means that RNNs seem to learn non-minimal representations of the concept that is being learned. OPTICS generally improves upon DBSCAN, but in this use case it seems to perform 
slightly worse than DBSCAN. The accuracy of clusters computed by mean shift depends on the parameterization of the bandwidth $bw$. The bandwidth estimated by scikit-learn results in a number of clusters that is smaller than $n$, causing high ambiguity. But as we decrease $bw$, in turn increasing the number of found clusters, mean shift becomes the best clustering method, even outperforming supervised classification approaches like LDA and \gls*{LR}. Regarding \textbf{RQ2}, we can state that \emph{clusters correlate with automaton states}, with the caveat that proper 
parameterization is necessary.

Detailed results with mean and standard deviation, as well as maximum ambiguity and number of perfect clusterings of each approach are in the appendix. %in the Table~\ref{tab:normal_training_table}.

\subsection{Impact of Network Architecture}
Next, we examine the influence of the 
RNN architecture on classification and clustering.
Figure~\ref{fig:different_rnn_types} shows the
average weighted ambiguity achieved by \gls*{LR}  
and selected clustering parameterizations
for each RNN architecture separately. 
GRU networks appear to create state space 
structures that lend themselves best to clustering
with any of the approaches. Interestingly, 
Elman ReLU RNNs lead to the highest ambiguity
on average, even for \gls*{LR}. A potential 
explanation is that
ReLU activations are unbounded, whereas $\tanh$,
e.g., is bounded. Hence, $\tanh$ Elman
RNNs may create clusters in the saturated area 
of $\tanh$, as in our RNN encoding of automata. 
We conclude from these experiments that clustering
techniques should be chosen in accordance with
the RNN type under consideration.

\begin{figure}[t]
    \centering
    \input{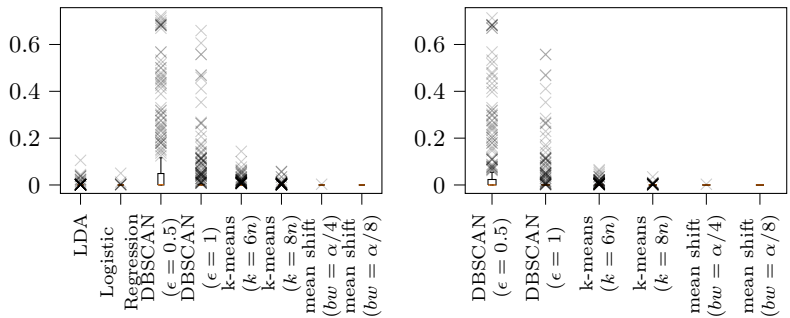}
    % This file was created with tikzplotlib v0.10.1.
\begin{tikzpicture}

\definecolor{darkgray176}{RGB}{176,176,176}
\definecolor{darkorange25512714}{RGB}{255,127,14}

\begin{axis}[
tick align=outside,
tick pos=left,
x grid style={darkgray176},
xmin=0.5, xmax=6.5,
xtick style={color=black},
x tick label style={rotate=90,anchor=east,font=\scriptsize},
y grid style={darkgray176},
ymin=-0.0360593717138064, ymax=0.757246805989935,
ytick style={color=black},
xtick={1,2,3,4,5,6},
xticklabels={\shortstack{DBSCAN \\ ($\epsilon=0.5$)}, \shortstack{DBSCAN \\ ($\epsilon=1$)}, \shortstack{k-means \\ ($k=6n$)}, \shortstack{k-means \\ ($k=8n$)}, \shortstack{mean shift \\ ($bw = \alpha /4$)}, \shortstack{mean shift \\ ($bw = \alpha /8$)}},
width = 0.48\textwidth,
height= .21\textheight,
]
\addplot [black]
table {%
0.925 0
1.075 0
1.075 0.023387503989235
0.925 0.023387503989235
0.925 0
};
\addplot [black]
table {%
1 0
1 0
};
\addplot [black]
table {%
1 0.023387503989235
1 0.0543366865878877
};
\addplot [black]
table {%
0.9625 0
1.0375 0
};
\addplot [black]
table {%
0.9625 0.0543366865878877
1.0375 0.0543366865878877
};
\addplot [black, mark=x, opacity=0.2, mark size=3, mark options={solid,fill opacity=0}, only marks]
table {%
1 0.329931346507478
1 0.273945065658385
1 0.319105224283193
1 0.397695785434508
1 0.681667259736663
1 0.679726381296272
1 0.683726949391589
1 0.567818518124529
1 0.533168152520805
1 0.670306926133925
1 0.669465792375921
1 0.682437806130453
1 0.6848155268084
1 0.0660360966890104
1 0.0783107485630174
1 0.115974983030594
1 0.297335842762734
1 0.518003165281061
1 0.479732620474054
1 0.202308666549685
1 0.213926131477257
1 0.163404881529087
1 0.06914184267461
1 0.0617703878423624
1 0.713240790749507
1 0.232949780559359
1 0.295912024839989
1 0.0846100688858122
1 0.108434901380383
1 0.0877123596151377
1 0.239790887754595
1 0.259429052144437
1 0.076615490186044
1 0.146282368565347
1 0.179235871108201
1 0.302608227562123
1 0.316679048471315
1 0.0800170514227054
1 0.276205456617659
1 0.253188140491718
1 0.0724362547700769
1 0.154881623762012
1 0.108618402849859
1 0.454918326530436
1 0.472603570741393
1 0.10876555643189
1 0.179308564542102
1 0.413193135793524
1 0.433503513140088
1 0.0654474635656942
1 0.0754474544254627
1 0.191171380944889
1 0.110658186094188
1 0.202864281352129
1 0.246913706421724
};
\addplot [black]
table {%
1.925 0
2.075 0
2.075 0.000554312249064382
1.925 0.000554312249064382
1.925 0
};
\addplot [black]
table {%
2 0
2 0
};
\addplot [black]
table {%
2 0.000554312249064382
2 0.0012614826292847
};
\addplot [black]
table {%
1.9625 0
2.0375 0
};
\addplot [black]
table {%
1.9625 0.0012614826292847
2.0375 0.0012614826292847
};
\addplot [black, mark=x, opacity=0.2, mark size=3, mark options={solid,fill opacity=0}, only marks]
table {%
2 0.117165857510625
2 0.118263621528923
2 0.112789007841764
2 0.0989472131870096
2 0.0617063803204666
2 0.262636241775564
2 0.11237071678771
2 0.11365325918169
2 0.111067695971373
2 0.109386451481698
2 0.0295487119111687
2 0.038347357076105
2 0.0294156781158495
2 0.0450266382618293
2 0.0743344633321044
2 0.171256734104658
2 0.263045159461444
2 0.082878026483392
2 0.207353929040339
2 0.352949117358473
2 0.467553319547431
2 0.557995385668788
2 0.0149905588948526
2 0.0143976829358182
2 0.284248106904106
2 0.267863187440455
2 0.470117423863287
2 0.411170788563956
2 0.0429958379119468
2 0.229012935354975
2 0.0502563494741996
2 0.0445431965804675
2 0.138905850439595
2 0.159190147678626
2 0.0014504472506922
2 0.00233297876420074
2 0.556747213017946
2 0.048522649268316
2 0.00875677587189856
2 0.0935493128581005
2 0.00140574105913967
2 0.00162011696254734
2 0.0214136035337922
2 0.038478796538222
2 0.0391772843360204
2 0.0524731390794539
2 0.00315611976287291
2 0.0151635281295434
2 0.00756768252969112
2 0.00284224596087897
2 0.0801763822040116
2 0.0924891875990071
2 0.00942345514872729
2 0.0530117998035087
2 0.0512392310232347
2 0.00494595618470022
2 0.00833041492221497
2 0.0277939673551415
2 0.049132167783342
};
\addplot [black]
table {%
2.925 0
3.075 0
3.075 0
2.925 0
2.925 0
};
\addplot [black]
table {%
3 0
3 0
};
\addplot [black]
table {%
3 0
3 0
};
\addplot [black]
table {%
2.9625 0
3.0375 0
};
\addplot [black]
table {%
2.9625 0
3.0375 0
};
\addplot [black, mark=x, opacity=0.2, mark size=3, mark options={solid,fill opacity=0}, only marks]
table {%
3 0.00250723811199555
3 0.00346893646578087
3 0.00577311417670525
3 0.00637746776577506
3 0.0115443402387035
3 0.00230242343695443
3 0.00908182769306116
3 0.000485067521640295
3 0.0518192571713356
3 0.00967621302713456
3 0.0210448264094866
3 0.0144796529332155
3 0.0174441430097322
3 0.0216135880568956
3 0.0199060374984035
3 0.00648714229490362
3 0.00578916337594279
3 0.00279676600700626
3 0.0218267464502719
3 0.0107187429569861
3 0.00496548446542576
3 0.0207455335927473
3 0.00330731251644706
3 0.00903650591814466
3 0.0371477020879904
3 0.0595519150521465
3 0.0128871675981708
3 0.0172824011384017
3 0.0348485920987399
3 0.0155960484269263
3 0.0116175267638647
3 0.0451375017881605
3 0.0167031212106602
3 0.0340382127224637
3 0.06509367822813
3 0.00746626703124639
3 0.00811880792580755
3 0.0179205900263307
3 0.0229149384231934
3 0.000571044591211288
3 0.00109435512897616
3 0.00482748060385513
3 0.000970575715367183
3 0.00613110087254534
3 0.00323341133085461
3 0.00599539978628563
3 0.006081861111608
3 0.00279161696709053
3 0.00752084254813429
3 0.00499623139200675
3 0.0022898791643248
3 0.00415851948892418
3 0.00323645823207159
3 0.00293411574282752
3 0.00738375434280842
3 0.00430623258601747
};
\addplot [black]
table {%
3.925 0
4.075 0
4.075 0
3.925 0
3.925 0
};
\addplot [black]
table {%
4 0
4 0
};
\addplot [black]
table {%
4 0
4 0
};
\addplot [black]
table {%
3.9625 0
4.0375 0
};
\addplot [black]
table {%
3.9625 0
4.0375 0
};
\addplot [black, mark=x, opacity=0.2, mark size=3, mark options={solid,fill opacity=0}, only marks]
table {%
4 0.00173063080331506
4 0.00315715538992126
4 0.00299718957116828
4 0.000485067521640295
4 0.0344850928741454
4 0.00741440066178696
4 0.000257291807179583
4 0.00703681958198756
4 0.00646281374312525
4 0.00350280702554916
4 0.00472256580770093
4 0.00384747490714516
4 0.00380942721338775
4 0.00976185927583595
4 0.000428465637866686
4 0.000420278695989226
4 0.00339261412035345
4 0.00700908034538047
4 0.00609169361601089
4 0.0105275747081911
4 0.00376192566207193
4 0.00227333055988598
4 0.000449211957558264
4 0.00396177256666251
4 0.00144845488206131
4 0.000744008807318861
};
\addplot [black]
table {%
4.925 0
5.075 0
5.075 0
4.925 0
4.925 0
};
\addplot [black]
table {%
5 0
5 0
};
\addplot [black]
table {%
5 0
5 0
};
\addplot [black]
table {%
4.9625 0
5.0375 0
};
\addplot [black]
table {%
4.9625 0
5.0375 0
};
\addplot [black, mark=x, opacity=0.2, mark size=3, mark options={solid,fill opacity=0}, only marks]
table {%
5 0.00303606154301966
};
\addplot [black]
table {%
5.925 0
6.075 0
6.075 0
5.925 0
5.925 0
};
\addplot [black]
table {%
6 0
6 0
};
\addplot [black]
table {%
6 0
6 0
};
\addplot [black]
table {%
5.9625 0
6.0375 0
};
\addplot [black]
table {%
5.9625 0
6.0375 0
};
\addplot [darkorange25512714]
table {%
0.925 0
1.075 0
};
\addplot [darkorange25512714]
table {%
1.925 0
2.075 0
};
\addplot [darkorange25512714]
table {%
2.925 0
3.075 0
};
\addplot [darkorange25512714]
table {%
3.925 0
4.075 0
};
\addplot [darkorange25512714]
table {%
4.925 0
5.075 0
};
\addplot [darkorange25512714]
table {%
5.925 0
6.075 0
};
\end{axis}

\end{tikzpicture}
    \caption{Boxplots of the weighted ambiguity resulting from different 
    clustering techniques for all $313$ experiments with GRUs (left) and  
    for a subset of $278$ GRU experiments, where LDA and \gls*{LR} acheived perfect separability.}
    \vspace{-0.3cm}
    \label{fig:cluster_ambiguity_grus}
\end{figure}
\noindent
\textbf{Clustering of GRU Networks.}
In the following, we examine the clustering hypothesis for GRU
networks, for which we found particularly low 
ambiguity measurements. Figure~\ref{fig:cluster_ambiguity_grus}
shows box plots of ambiguity values measured for different 
clustering techniques. On the left, we show measurements
from all experiments involving GRUs and on the right, we
show measurements from experiments, where LDA is able to perfectly
separate hidden states corresponding to all automaton states.

We found that the hidden state space of the trained GRUs has a structure that is well-suited for linear separation
of states. In $88\%$ of the experiments, both LDA and \gls*{LR} can perfectly separate states. 

This benefits k-means as well, which achieves 
lower ambiguity in all considered parameterizations when compared to other network types. 
In particular, k-means with $k=8n$ performs very well, where the third quartile of
the ambiguity values is equal to zero.  From the other techniques, 
DBSCAN with $\epsilon=1$ benefits from restricting the analysis
to GRUs. OPTICS and especially mean shift are hardly affected.
Focusing only on the experiments where LDA has an ambiguity of
$0$ (Fig.~\ref{fig:cluster_ambiguity_grus} (right)), we see that
k-means performs especially well, with an ambiguity of at most
$0.034$ for $k=8n$.

\subsection{Training of Noisy Correct-by-Construction RNNs}

The second set of experiments considers trained RNNs that have been initialized to the noisy ``correct-by-construction'' RNNs. That is, for each DFA we have computed an RNN with the method discussed in Sect.~\ref{sec:rnn_construction} and introduced Gaussian noise over the weights. The training was then performed on such an RNN initialized with noisy close-to-perfect weights. 
%A visualization of the network's hidden state space at each stage of this process can be seen in Fig.~\ref{fig:retraining_of_constructed_RNN}. 
This set of experiments was performed on all DFAs, with a single RNN size and type, and three configurations of %per regular language with varying 
saturation and noise parameters. We considered only RNNs that achieved 100\% accuracy, thus reducing this set of experiments to 95 RNNs.

\begin{figure}[t]
    \centering
    \input{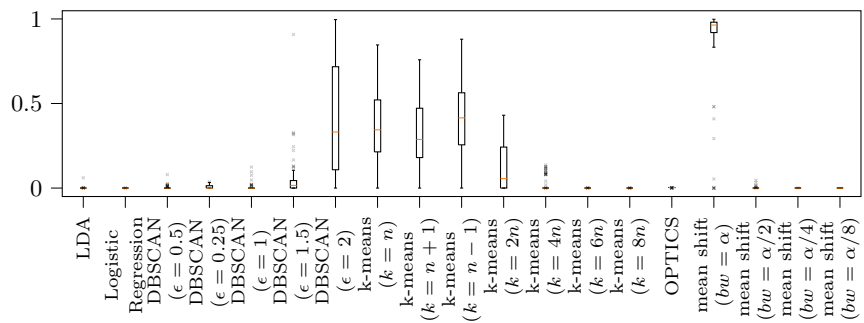}
    \vspace{-0.3cm}
    \caption{Boxplots of the clustering ambiguity resulting from different
    clustering techniques from 95 trained noisy constructed RNNs.}
    \vspace{-0.3cm}
    \label{fig:clustering_retrained_perfect_acc}
\end{figure}

The results can be seen in Fig.~\ref{fig:clustering_retrained_perfect_acc} and in the third row of Table~\ref{tab:number_of_perfect_clustering}. Both LDA and \gls*{LR} were able to perfectly classify hidden-state vectors in 92\% and 95\% percent of the cases, respectively. DBSCAN managed to compute clusters with low ambiguity, but compared to the previous set of experiments, it achieved so with smaller $\epsilon$. On average, OPTICS found clusters with very low ambiguity (0.003 $\pm$ 0.0008), but interestingly it found perfect clusterings in only 17\% of the cases. K-means had similar performance as in the previous set of experiments, with $k=6n$ and $k=8n$ achieving near-perfect results. Once again, mean shift outperformed both LDA and \gls*{LR}, as well as other unsupervised clustering approaches.
Concerning \textbf{RQ3}, we can state that \emph{initial conditions do affect clustering}. For example, our constructed weights seem to lead to smaller distances in the state space, such that DBSCAN performed better with smaller $\epsilon$.

\begin{table}[t]
\vspace{-0.3cm}
\caption{Number of perfect clustering (zero ambiguity) achieved by selected clustering method over both set of experiments.}
\vspace{-0.3cm}
\begin{center}
{ \scriptsize
\begin{tabular}{|l|r|r|r|r|r|r|r|r|r|r|r|r|}
\hline
\begin{tabular}[c]{@{}l@{}}Clustering Function\end{tabular} &
  \multicolumn{1}{l|}{LDA} &
  \multicolumn{1}{|c|}{LR} &
  \multicolumn{3}{|c|}{
  DBSCAN
  }
  &
%   \multicolumn{1}{l|}{\begin{tabular}[c]{@{}l@{}}k-means\\ ($k=n$)\end{tabular}} &
%   \multicolumn{1}{l|}{\begin{tabular}[c]{@{}l@{}}k-means\\ ($k=6n$)\end{tabular}} &
%   \multicolumn{1}{l|}{\begin{tabular}[c]{@{}l@{}}k-means\\ ($k=8n$)\end{tabular}} &
  \multicolumn{3}{|c|}{
  k-means
  }
  &
  \multicolumn{1}{l|}{OPTICS} &
  \multicolumn{3}{|c|}{
  mean shift
  }
  \\\hline
  Parameters 
  & % LDA
  & % LR
  & $0.5$ & $0.25$ & $1.5$ 
  & $n$ & $6n$ & $8n$ 
  & %OPTICS
  & $\alpha/2$ & $\alpha/4$ & $\alpha/8$ \\\hline
%   \multicolumn{1}{l|}{\begin{tabular}[c]{@{}l@{}}mean shift\\ ($bw=\alpha/2$)\end{tabular}} &
%   \multicolumn{1}{l|}{\begin{tabular}[c]{@{}l@{}}mean shift\\ ($bw=\alpha/4$)\end{tabular}} &
%   \multicolumn{1}{l|}{\begin{tabular}[c]{@{}l@{}}mean shift\\ ($bw=\alpha/8$)\end{tabular}} \\ 
\begin{tabular}[c]{@{}l@{}}\# Perfect Clustering\\ (1350 experiments)\end{tabular} &
  1003 &
  1235 &
  368 &
  185 &
  639 &
  49 &
  629 &
  783 &
  16 &
  373 &
  1296 &
  1331 \\ \hline
\begin{tabular}[c]{@{}l@{}}\# Perfect Clustering\\ (95 Experiments)\end{tabular} &
  88 &
  92 &
  80 &
  51 &
  42 &
  3 &
  90 &
  90 &
  3 &
  85 &
  93 &
  74 \\ \hline
\end{tabular}%
}
\end{center}
\vspace{-0.6cm}
\label{tab:number_of_perfect_clustering}
\end{table}

\subsection{Number of Clusters}\label{sec:cluster_sizes}

\begin{figure}[t]
    \centering
    % This file was created with tikzplotlib v0.10.1.
\begin{tikzpicture}

\definecolor{darkgray176}{RGB}{176,176,176}
\definecolor{darkorange25512714}{RGB}{255,127,14}

\begin{axis}[
tick align=outside,
tick pos=left,
x grid style={darkgray176},
xmin=0.5, xmax=8.5,
xtick style={color=black},
x tick label style={rotate=90,anchor=east,font=\scriptsize},
y grid style={darkgray176},
ymin=-70.3, ymax=1520.3,
ytick style={color=black},
xtick={1,2,3,4,5,6,7,8},
xticklabels={LDA, \shortstack{Logistic\\ Regression}, \shortstack{DBSCAN\\ ($\epsilon=0.5$)}, \shortstack{DBSCAN\\ ($\epsilon=1$)},\shortstack{k-means\\ ($k=6n$)},\shortstack{k-means\\ ($k=8n$)}, \shortstack{mean shift\\ ($bw = \alpha /4$)}, \shortstack{mean shift\\ ($bw = \alpha /8$)}},
width = .5\textwidth,
height= .21\textheight,
]
\addplot [black]
table {%
0.925 5
1.075 5
1.075 10
0.925 10
0.925 5
};
\addplot [black]
table {%
1 5
1 5
};
\addplot [black]
table {%
1 10
1 12
};
\addplot [black]
table {%
0.9625 5
1.0375 5
};
\addplot [black]
table {%
0.9625 12
1.0375 12
};
\addplot [black]
table {%
1.925 5
2.075 5
2.075 10
1.925 10
1.925 5
};
\addplot [black]
table {%
2 5
2 5
};
\addplot [black]
table {%
2 10
2 12
};
\addplot [black]
table {%
1.9625 5
2.0375 5
};
\addplot [black]
table {%
1.9625 12
2.0375 12
};
\addplot [black]
table {%
2.925 74.25
3.075 74.25
3.075 325.5
2.925 325.5
2.925 74.25
};
\addplot [black]
table {%
3 74.25
3 2
};
\addplot [black]
table {%
3 325.5
3 531
};
\addplot [black]
table {%
2.9625 2
3.0375 2
};
\addplot [black]
table {%
2.9625 531
3.0375 531
};
\addplot [black]
table {%
3.925 35
4.075 35
4.075 222
3.925 222
3.925 35
};
\addplot [black]
table {%
4 35
4 2
};
\addplot [black]
table {%
4 222
4 500
};
\addplot [black]
table {%
3.9625 2
4.0375 2
};
\addplot [black]
table {%
3.9625 500
4.0375 500
};
\addplot [black, mark=x, opacity=0.2, opacity=0.2, opacity=0.2, mark size=0.8, mark options={solid,fill opacity=0}, only marks]
table {%
4 506
4 535
4 542
4 539
4 550
4 508
4 547
4 536
4 515
4 518
4 568
4 538
};
\addplot [black]
table {%
4.925 30
5.075 30
5.075 60
4.925 60
4.925 30
};
\addplot [black]
table {%
5 30
5 30
};
\addplot [black]
table {%
5 60
5 72
};
\addplot [black]
table {%
4.9625 30
5.0375 30
};
\addplot [black]
table {%
4.9625 72
5.0375 72
};
\addplot [black]
table {%
5.925 40
6.075 40
6.075 80
5.925 80
5.925 40
};
\addplot [black]
table {%
6 40
6 40
};
\addplot [black]
table {%
6 80
6 96
};
\addplot [black]
table {%
5.9625 40
6.0375 40
};
\addplot [black]
table {%
5.9625 96
6.0375 96
};
\addplot [black]
table {%
6.925 36.25
7.075 36.25
7.075 166.75
6.925 166.75
6.925 36.25
};
\addplot [black]
table {%
7 36.25
7 4
};
\addplot [black]
table {%
7 166.75
7 361
};
\addplot [black]
table {%
6.9625 4
7.0375 4
};
\addplot [black]
table {%
6.9625 361
7.0375 361
};
\addplot [black, mark=x, opacity=0.2, opacity=0.2, mark size=0.8, mark options={solid,fill opacity=0}, only marks]
table {%
7 1052
7 603
7 610
7 723
7 435
7 900
7 1293
7 1139
7 1373
7 1277
7 1388
7 1425
7 1373
7 831
7 930
7 1053
7 842
7 462
7 455
7 477
7 643
7 736
7 709
7 774
7 693
7 696
7 652
7 408
7 603
7 614
7 885
7 971
7 803
7 749
7 1019
7 889
7 454
7 468
7 373
7 610
7 583
7 617
7 661
7 441
7 634
7 403
7 425
7 475
7 417
7 744
7 720
7 625
7 622
7 773
7 769
7 447
7 505
7 452
7 487
7 435
7 372
7 367
7 606
7 404
7 684
7 691
7 1261
7 847
7 702
7 598
7 394
7 571
7 589
7 829
7 934
7 715
7 553
7 563
7 458
7 684
7 737
7 465
7 610
7 468
7 377
7 382
7 599
7 406
7 522
7 371
7 425
};
\addplot [black]
table {%
7.925 85
8.075 85
8.075 630.75
7.925 630.75
7.925 85
};
\addplot [black]
table {%
8 85
8 9
};
\addplot [black]
table {%
8 630.75
8 1448
};
\addplot [black]
table {%
7.9625 9
8.0375 9
};
\addplot [black]
table {%
7.9625 1448
8.0375 1448
};
\addplot [darkorange25512714]
table {%
0.925 8
1.075 8
};
\addplot [darkorange25512714]
table {%
1.925 8
2.075 8
};
\addplot [darkorange25512714]
table {%
2.925 173
3.075 173
};
\addplot [darkorange25512714]
table {%
3.925 93
4.075 93
};
\addplot [darkorange25512714]
table {%
4.925 48
5.075 48
};
\addplot [darkorange25512714]
table {%
5.925 64
6.075 64
};
\addplot [darkorange25512714]
table {%
6.925 76
7.075 76
};
\addplot [darkorange25512714]
table {%
7.925 231
8.075 231
};
\end{axis}

\end{tikzpicture}
        % This file was created with tikzplotlib v0.10.1.
\begin{tikzpicture}

\definecolor{darkgray176}{RGB}{176,176,176}
\definecolor{darkorange25512714}{RGB}{255,127,14}

\begin{axis}[
tick align=outside,
tick pos=left,
x grid style={darkgray176},
xmin=0.5, xmax=8.5,
xtick style={color=black},
x tick label style={rotate=90,anchor=east,font=\scriptsize},
y grid style={darkgray176},
ymin=-4.45, ymax=159.45,
ytick style={color=black},
xtick={1,2,3,4,5,6,7,8},
xticklabels={LDA, \shortstack{Logistic\\ Regression}, \shortstack{DBSCAN\\ ($\epsilon=0.5$)}, \shortstack{DBSCAN\\ ($\epsilon=1$)},\shortstack{k-means\\ ($k=6n$)},\shortstack{k-means\\ ($k=8n$)}, \shortstack{mean shift\\ ($bw = \alpha /4$)}, \shortstack{mean shift\\ ($bw = \alpha /8$)}},
width = .5\textwidth,
height= .21\textheight,
]
\addplot [black]
table {%
0.925 5
1.075 5
1.075 10
0.925 10
0.925 5
};
\addplot [black]
table {%
1 5
1 4
};
\addplot [black]
table {%
1 10
1 10
};
\addplot [black]
table {%
0.9625 4
1.0375 4
};
\addplot [black]
table {%
0.9625 10
1.0375 10
};
\addplot [black]
table {%
1.925 5
2.075 5
2.075 10
1.925 10
1.925 5
};
\addplot [black]
table {%
2 5
2 4
};
\addplot [black]
table {%
2 10
2 10
};
\addplot [black]
table {%
1.9625 4
2.0375 4
};
\addplot [black]
table {%
1.9625 10
2.0375 10
};
\addplot [black]
table {%
2.925 10
3.075 10
3.075 31
2.925 31
2.925 10
};
\addplot [black]
table {%
3 10
3 3
};
\addplot [black]
table {%
3 31
3 61
};
\addplot [black]
table {%
2.9625 3
3.0375 3
};
\addplot [black]
table {%
2.9625 61
3.0375 61
};
\addplot [black]
table {%
3.925 9.25
4.075 9.25
4.075 30
3.925 30
3.925 9.25
};
\addplot [black]
table {%
4 9.25
4 3
};
\addplot [black]
table {%
4 30
4 60
};
\addplot [black]
table {%
3.9625 3
4.0375 3
};
\addplot [black]
table {%
3.9625 60
4.0375 60
};
\addplot [black]
table {%
4.925 30
5.075 30
5.075 60
4.925 60
4.925 30
};
\addplot [black]
table {%
5 30
5 24
};
\addplot [black]
table {%
5 60
5 60
};
\addplot [black]
table {%
4.9625 24
5.0375 24
};
\addplot [black]
table {%
4.9625 60
5.0375 60
};
\addplot [black]
table {%
5.925 40
6.075 40
6.075 80
5.925 80
5.925 40
};
\addplot [black]
table {%
6 40
6 32
};
\addplot [black]
table {%
6 80
6 80
};
\addplot [black]
table {%
5.9625 32
6.0375 32
};
\addplot [black]
table {%
5.9625 80
6.0375 80
};
\addplot [black]
table {%
6.925 12.25
7.075 12.25
7.075 33.25
6.925 33.25
6.925 12.25
};
\addplot [black]
table {%
7 12.25
7 8
};
\addplot [black]
table {%
7 33.25
7 62
};
\addplot [black]
table {%
6.9625 8
7.0375 8
};
\addplot [black]
table {%
6.9625 62
7.0375 62
};
\addplot [black]
table {%
7.925 18.25
8.075 18.25
8.075 50
7.925 50
7.925 18.25
};
\addplot [black]
table {%
8 18.25
8 10
};
\addplot [black]
table {%
8 50
8 90
};
\addplot [black]
table {%
7.9625 10
8.0375 10
};
\addplot [black]
table {%
7.9625 90
8.0375 90
};
\addplot [black, mark=x, opacity=0.2, mark size=0.8, mark options={solid,fill opacity=0}, only marks]
table {%
8 152
8 112
8 111
8 103
8 129
};
\addplot [darkorange25512714]
table {%
0.925 5
1.075 5
};
\addplot [darkorange25512714]
table {%
1.925 5
2.075 5
};
\addplot [darkorange25512714]
table {%
2.925 21
3.075 21
};
\addplot [darkorange25512714]
table {%
3.925 19.5
4.075 19.5
};
\addplot [darkorange25512714]
table {%
4.925 30
5.075 30
};
\addplot [darkorange25512714]
table {%
5.925 40
6.075 40
};
\addplot [darkorange25512714]
table {%
6.925 20.5
7.075 20.5
};
\addplot [darkorange25512714]
table {%
7.925 30.5
8.075 30.5
};
\end{axis}

\end{tikzpicture}
        \vspace{-0.3cm}
    \caption{Number of clusters found in experiments
    with RNNs trained from scratch (left) and with RNNs initialized with close-to-perfect weights (right).
    }
    % with ambiguity scores depicted in Fig.~\ref{fig:clustering_all_80_percent_acc} (left) and Fig.~\ref{fig:clustering_retrained_perfect_acc} (right).}
    \vspace{-0.65cm}
    \label{fig:cluster_size_overview}
\end{figure}
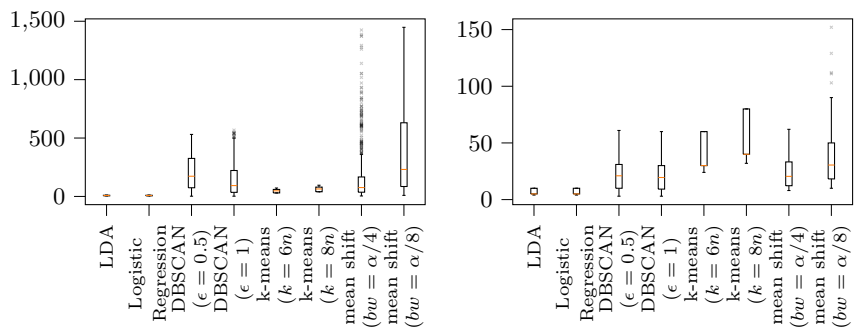

% At a first glance, we
% may conclude that the clustering hypothesis does not hold in general. 
% However, in a large number of cases, the clustering occurs  when the  
% \emph{correct} approach is chosen. The median weighted ambiguity
% of DBSCAN with $\epsilon\geq1$ is almost equal to $0$, which means that
% in half of the cases, we can identify semantically states from clusters.
% Likewise, k-means achieves a low ambiguity when given at least four times as many
% clusters as ``necessary''. This means that RNNs seem to learn non-minimal
% representations of the concept that is being learned. 
% We should contrast this
% with LDA, which is shown here as a reference. It achieves the lowest 
% ambiguity, which tells us that linear decision boundaries may be 
% sufficient in most cases. In fact, LDA separates the hidden state 
% space into $n$ regions,
% thus k-means with $k=n$ is potentially sufficient. Recall, however, that
% LDA is derived using additional information about the correspondence between
% hidden and automaton states, which is usually not available.
% OPTICS generally improves upon DBSCAN, but in this use case it seems to perform 
% slightly worse than DBSCAN, so we will concentrate on DBSCAN in the following.
% Mean shift in its default parameterization fails to meaningful clusters in most cases.
% Finally, we can see that the best-performing approaches
% have many outliers in terms of ambiguity. Hence, 
% without additional knowledge
% we should not rely on the clustering hypothesis, since all approaches may
% perform poorly.

We only considered ambiguity so far. As clustering is a tool for 
abstraction, we also need to look at the size of the abstraction, i.e., 
the number of clusters. 
K-means fixes this size, but the number 
of clusters derived by other approaches depends on
their parameters \emph{and} the given data. Figure~\ref{fig:cluster_size_overview} shows 
box plots of the number of clusters derived by different techniques for both previously discussed experiments. We show two instantiations of k-means as a reference. 

We observe interesting trends between the two experiments. From Fig.~\ref{fig:cluster_size_overview}, we can observe that the accuracy and initialization of an RNN has a large impact on the number of clusters. The right subplot of Fig.~\ref{fig:cluster_size_overview} shows that both DBSCAN parameterizations yield smaller numbers of clusters than k-means, but in the case where RNNs are not perfectly accurate ($\geq 80\%$)
and are trained from scratch, DBSCAN finds substantially more clusters than k-means. The same trend, but with even bigger size differences can be observed for mean shift. These findings imply that while clustering functions with low ambiguity can be found for non-perfect RNNs, they tend to have more clusters than clustering functions computed on perfectly accurate RNNs. The fact that mean shift with $bw=\alpha/8$ finds up to $1.5k$
clusters in $\sim 2.5k$ data points (recall that we reduced the data for mean shift) and that the third quartile in Fig.~\ref{fig:cluster_size_overview} (left) is $500$ weakens our findings regarding \textbf{RQ2}. The highly accurate clusterings found with mean shift hardly group states in many cases, thus making it trivial to achieve high accuracy. On average, most clustering approaches would improve the efficiency of model-based analysis of RNNs by reducing approximately $10k$ data points in the validation data to a few hundred clusters. 

\noindent
\setlength{\intextsep}{0pt}%

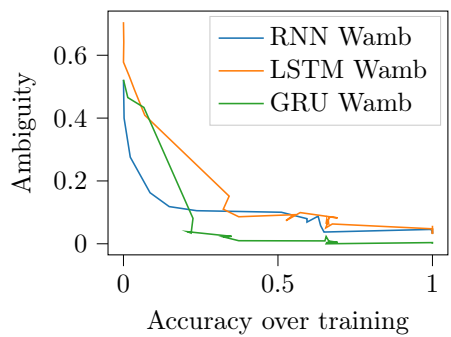
\begin{wrapfigure}[14]{t}{.5\textwidth}
  \centering
  % This file was created with tikzplotlib v0.10.1.
\begin{tikzpicture}

\definecolor{darkgray176}{RGB}{176,176,176}
\definecolor{darkorange25512714}{RGB}{255,127,14}
\definecolor{forestgreen4416044}{RGB}{44,160,44}
\definecolor{lightgray204}{RGB}{204,204,204}
\definecolor{steelblue31119180}{RGB}{31,119,180}

\begin{axis}[
legend cell align={left},
legend style={fill opacity=0.8, draw opacity=1, text opacity=1, draw=lightgray204},
tick align=outside,
tick pos=left,
x grid style={darkgray176},
xlabel={Accuracy over training},
xmin=-0.05, xmax=1.05,
xtick style={color=black},
y grid style={darkgray176},
ylabel={Ambiguity},
ymin=-0.0352684012986692, ymax=0.740636427272054,
ytick style={color=black},
width = .5\textwidth,
height= .25\textheight,
]
\addplot [semithick, steelblue31119180]
table {%
0 0.521876612335541
0.002 0.39962560736377
0.022 0.275689922265469
0.086 0.162470533955105
0.148 0.118327125729326
0.237 0.105269871375797
0.511 0.100317519588481
0.594 0.0798101022557862
0.594 0.0689810900400837
0.63 0.0878336897956714
0.638 0.0578892049283314
0.649 0.0373516251671254
0.991 0.0456199947554358
0.999 0.0441093925456847
1 0.0326923085554083
1 0.0328048124218249
1 0.0477638100714078
};
\addlegendentry{RNN Wamb}
\addplot [semithick, darkorange25512714]
table {%
0 0.705368025973385
0.001 0.641915366587167
0 0.578014515698116
0.019 0.532898425436535
0.036 0.489017413610913
0.069 0.409111373212271
0.342 0.151582759467149
0.323 0.109966470852831
0.373 0.0859866645480311
0.551 0.0924745301830232
0.526 0.0736517771734062
0.572 0.0992590288737955
0.693 0.0824282554410868
0.668 0.085351992426817
0.66 0.0705793401251532
0.658 0.0782586449164707
0.656 0.0654995486009894
0.664 0.0631278692511474
0.663 0.0741526758813666
0.655 0.0506773857110169
0.676 0.0629985458931364
1 0.0476302468211591
1 0.0309228916897423
1 0.0573345879844228
};
\addlegendentry{LSTM Wamb}
\addplot [semithick, forestgreen4416044]
table {%
0 0.521875018033389
0.014 0.465339526695216
0.0659999999999999 0.433971501770644
0.225 0.0809251350202236
0.219 0.0339585772268272
0.207 0.0384343437276459
0.35 0.0236804473774296
0.323 0.0220591401485422
0.373 0.00975154622544929
0.652 0.00918809576563958
0.656 0.0230423604795519
0.667 0.00452627567951132
0.693 0.00452627567951132
0.668 0.00834993695287054
0.66 0.00376256948954344
0.658 0.00376256948954344
0.656 0
1 0.00376256948954344
1 0
1 0
};
\addlegendentry{GRU Wamb}
\end{axis}

\end{tikzpicture}
  \vspace{-0.6cm}
  \caption{Relationship between accuracy and ambiguity over the training process.}
  \label{fig:accuract_ambiguity_relation}
\end{wrapfigure}
\emph{Discussion.} Given our findings on the number of clusters, we should consider techniques and parameterizations that create reasonably-sized abstractions. K-means with $k=8n$ achieves perfect clustering in $58\%$ of the cases on RNNs trained from scratch (see Table~\ref{tab:number_of_perfect_clustering}). Additionally, we found that this k-means parameterization achieved a mean $\mathit{wamb}$ of $0.018$ overall, but in the most extreme case, $\mathit{wamb}$ was $0.67$. Hence, clusters mostly correlate with automaton states, if an appropriate clustering technique is applied. As such reasonably-sized clustering functions do not achieve high accuracy in some cases, we cannot rely on the clustering hypothesis alone in safety-critical applications, like autonomous driving. Additional techniques are needed to mitigate the error resulting from ambiguous clusters, like probabilistic automata learning~\cite{DBLP:conf/kbse/DongWSZWDDW20}.
To further validate the unambiguous (perfect) clustering functions, we extracted non-minimal automata from cluster labels, as described by Hou and Zhou~\cite{DBLP:journals/tnn/HouZ20}. Our findings conform to theirs: extracted non-minimal automata can be minimized to correspond to the ground-truth models.

%%%%%%%%%%%%%%%%%%%%

Figure~\ref{fig:accuract_ambiguity_relation} shows the relationship between RNN accuracy and ambiguity over the training process. We observe that the ambiguity decreases during the training process, as the accuracy increases. 
This shows that the clustering hypothesis is related 
to RNN accuracy, and even further validates the ambiguity metric introduced in Sect.~\ref{sec:ambiguity_computation}.
In only few cases, accuracy decreases slightly
as ambiguity decreases and vice versa,
which explains the kinks in the graphs. 
However, there is generally a strong negative 
correlation, i.e., the more accurate an RNN gets, the clustering gets less ambiguous: the Spearman correlation 
coefficients are $-0.87$, $-0.96$, and $-0.9$ 
for Elman RNNs, LSTMs, and GRUs in Fig.~\ref{fig:accuract_ambiguity_relation}. Additionally, if the RNN retains the accuracy over long sequences (100-200 input symbols), the clustering ambiguity remains constant.

\section{Conclusion and Future Work}\label{sec:conclusion}
% In this paper 
We revisited and formally analyzed the hypothesis that an RNN's hidden-state vectors tend to form clusters. In our analysis, we examined the clustering hypothesis for RNNs that were trained to recognize regular languages. This provided us with a ground truth that allowed us to compare the identified clusters with the original finite-state semantics that underlie the learned concept.
Additionally, we examined the (linear) separability of hidden-state vectors into regions corresponding to automata states. 

In our experiments, we observed that the \textbf{hidden-state vectors can be (piecewise linearly) separated} in most of the considered cases.  
This finding indicates that it is potentially possible to derive meaningful mappings from hidden-state vectors to automata states.
Considering unsupervised clustering, the regions identified by classifiers may contain multiple clusters and clusters may span across decision boundaries. 
However, we show that it is \textbf{possible to compute clustering functions that correlate with automata states}. For example, clusterings obtained by k-means with large $k$ overall achieved high accuracy and perfect accuracy in  $58\%$ of our experiments. Furthermore, in cases of perfect clustering, we observed that RNNs learn non-minimal representations of the regular languages on which they were trained.
The selection of a clustering method and its parameterization greatly affects the clustering accuracy, hindering the possibility of using an out-of-the-box clustering algorithm as a means of RNN hidden-state space abstraction. 

% In our experiments, we observed that the vectors can be linearly separated in most cases so we argue that it is potentially possible to derive a meaningful clustering. The regions identified by classifiers may contain several clusters though, and clusters may span across decision boundaries. We furthermore observed that clusters identified with unsupervised techniques often correspond to the states of minimal ground-truth automata, as measured by an entropy-based ambiguity metric. However, ambiguity is very sensitive to the parameterization of the clustering approach, and in order to be successful, it is necessary to choose a configuration s.t. sufficiently many clusters can be found. However, this is detrimental to applications of clustering as abstractions, where abstraction size affects efficiency of the subsequent RNN analysis.

%\todo{IP: did not understand the message between the lines you'd like to send here}.
%\todo{IP: in respect of the reading flow there seems to be something missing here that has been on your mind; I wouldn't see the following as a traditional summary, so maybe there's some glue missing} 

%Our findings suggest that (1) RNNs learn non-minimal representations, (2) initialization with additional knowledge helps the formation of meaningful clusters, and (3) that we currently \textbf{cannot rely on the clustering hypothesis}, since reasonably-sized clustering abstractions derived by k-means achieved perfect accuracy in only $58\%$ of our experiments\todo{IP: check sentence}.

We identify various avenues for future work. With RNN verification being our ultimate goal, we will investigate ways to mitigate imprecisions and errors resulting from imperfect clustering.
A stochastic interpretation of RNNs and stochastic automata learning over clusters seems 
promising~\cite{DBLP:conf/kbse/DongWSZWDDW20}. %This could enable the construction of approximate finite-state representations of RNNs over potentially infinite input-output alphabets.
Further analysis of factors affecting the clustering is important as well, examples being the effect of different optimizers or overtraining.
Finally, we will investigate how to adapt the training and architecture of RNNs such that
clusters are likely to form, which would enable explainability-by-design. 

%needed to thoroughly analyze the relationship between found clusters and automaton states. We observed that hidden-state vectors are in most cases linearly separable, however, ... \todo{would say that CF had problems, but in general we can achieve low ambiguity}

%In feature work we will further examine the effect of various factors on the clustering hypothesis, such as increasing sequence length. Furthermore, we will examine possible application of presented methods on RNNs with high-dimensional input data-set. This could potentially enable us construction of approximate finite-state representations of potentially infinite RNN input-output alphabets. 

\paragraph{Acknowledgements.}
This work has been supported by the "University SAL Labs" initiative of Silicon Austria Labs (SAL) and its Austrian partner universities for applied fundamental research for electronic based systems.

\bibliographystyle{splncs04}
\bibliography{main}
\newpage
\appendix
\section{Appendix}
\subsection{Appendix A -- RNNs.}

\subsubsection{Mathematical definitions of RNNs.}

Below, we provide the mathematical definitions of LSTMs and GRUs following
the PyTorch implementation~\cite{NEURIPS2019_9015}.
\begin{align*}
\intertext{LSTMs:}
i_t &= \sigma(W_{ii} x_t + b_{ii} + W_{hi} h_{(t-1)} + b_{hi}) \\
f_t &= \sigma(W_{if} x_t + b_{if} + W_{hf} h_{(t-1)} + b_{hf}) \\
g_t &= \tanh(W_{ig} x_t + b_{ig} + r_t * ( W_{hg} h_{(t-1)} + b_{hg})) \\ 
o_t &= \sigma(W_{io} x_t + b_{io} + W_{ho} h_{(t-1)} + b_{ho}) \\
c_t &= f_t \odot c_{(t-1)} + i_t \odot g_t \text{ where } \odot \text{ is the Hadamard product and } \\&\sigma \text{ is the sigmoid function} \\
h_t &= o_t \odot \tanh(c_t)
\intertext{GRUs:}
r_t &= \sigma(W_{ir} x_t + b_{ir} + W_{hr} h_{(t-1)} + b_{hr}) \\
z_t &= \sigma(W_{iz} x_t + b_{iz} + W_{hz} h_{(t-1)} + b_{hz}) \\
n_t &= \tanh(W_{in} x_t + b_{in} + r_t \odot ( W_{hn} h_{(t-1)} + b_{hn})) \\ 
h_t &= (1-z_t) \odot n_t + z_t \odot h_{(t-1)}
\end{align*}

\subsubsection{Automata Encoding via RNNs.}

In the following, we provide a more detailed presentation of the 
RNN-based automata encoding. 
Recall that we use Elman RNNs with $\tanh$ activation and a single layer 
to encode automata. Similar encodings have been proposed in \cite{DBLP:journals/neco/AlquezarS95,DBLP:journals/tnn/GoudreauGCC94,minsky_book}.
%, therefore we keep the presentation brief. 

Our RNN-encoding of an automaton operates
on one-hot encoded transitions of the ground-truth DFA $A$. 
We encode transitions rather than states, because one-layer first-order RNNs
cannot encode all DFAs when they are restricted to one-hot encoded 
states~\cite{DBLP:journals/tnn/GoudreauGCC94}. 
Thus, we have for the hidden state space size $h = |Q| \cdot |\Sigma|$ and 
we impose a fixed ordering on the states $Q$ and the alphabet $\Sigma$. 
Let $t_{i,j} = (q_i, e_j)$ be the transition labeled $e_j$ originating in state
$q_i$. We denote its encoding $\enc(q_i,e_j)$, which is a vector in $\mathbb{R}^h$,
where exactly one element equals $1$ and all others are $-1$. 
The element set to $1$ is at the position $i + (j \cdot |Q|)$. Let $\psi(q_i)$
be the encoding of the $i^{th}$ state, 
a vector in  $\mathbb{R}^h$, which is $1$ at indexes
$i + j \cdot |Q|$ for $j\in \{1, \ldots, |\Sigma|\}$ and $-1$
otherwise. This encoding only shows up as a temporary value in the construction.
Furthermore, let $\enci(e_j)$ be the one-hot encoding of a symbol $e_j$,
a vector in $\mathbb{R}^{|\Sigma|}$
where exactly one element is equal to $1$ and all others are $0$.
The basic idea is to set the weights of an RNN $(r,o)$ so that $r$ always
operates in the saturated area of $\tanh$, making $\tanh$ work like a 
binary threshold gate. For this reason, $\enc$ and $\psi$ use $1$ and $-1$ (limits 
of $\tanh)$. At time step $t$,
the function $r$ maps the previous transition 
$\enc(q_{t-1}, e_{t-1})$ and the new input $\enci(e_t)$ to the 
encoding of the next transition $\enc(\delta(q_{t-1}, e_{t-1}),e_t)$,
where $\delta$ is the transition function of the DFA. 
To encode the whole DFA, we first set the RNN weights $W_{hh}$ such that it maps
an encoded transition to its encoded target state.
Hence, for each transition $(q_i,e_j)$, we need to 
ensure that $W_{hh} \pi(q_i,e_j) \approx \psi(\delta(q_i,e_j))$,
where $\delta(q_i,e_j)$ is the target state. As we apply $\tanh$ as activation, 
we formulate this as $W_{hh} \pi(q_i,e_j) \geq \psi(\delta(q_i,e_j))  H_r$, where $H_r$ is a
scalar factor defining how far we go into saturation. This factor ensures that 
we use $\tanh$ in a region, where it maps to values
 close to $-1$ and $1$.
For each of the $|Q|\cdot \Sigma|$ transitions, we have a system of $h = |Q|\cdot |\Sigma|$ linear inequations.
Hence, we have $h^2$ inequations and $h^2$ weights in the square matrix $W_{hh}$ 
that we can adjust. By considering that all transition encodings are unique, 
it is easy to show that the complete system of inequations always has a solution.

To map an encoded state and encoded input to an encoded transition, we set the remaining weights as follows:
\begin{align*}
    W_{ih} &= (wih)_{i=1,\ldots,|Q|\cdot|\Sigma|; j=1, \ldots,|\Sigma|}\text{, where } wih = 
    \begin{cases}
    0 \text{ if $\lfloor i / |\Sigma|  \rfloor = j$} \\
    -H_{r} \text{ otherwise}
    \end{cases} \\
    b_{ih} &= \mathbf{1} \cdot (-0.5) \cdot H_r \text{ and } b_{hh} = \mathbf{0}  \\
    &\text{ where } \mathbf{1} \text{ and } \mathbf{0} \text{ are vectors containing $1$ and $0$} 
    % T &= (t)_{i=1,\ldots,(|Q|\cdot|\Sigma|)^2; j=1, \ldots,(|Q|\cdot|\Sigma|)^2} \text{, where }\\
    % t &= 
    % \begin{cases}
    % \pi(q_k,e_l)_m \text{ if } i=k,l \\
    % 0 \text{ otherwise}
    % \end{cases} \\
    % \mathbf{s} &= (s)_{i=1,\ldots,|Q|\cdot|\Sigma|; j=1} \text{, where }\\
    % s &= \begin{cases}
    % x \text{ if $\lfloor i / |\Sigma|  \rfloor = j$} \\
    % 0 \text{ otherwise}
    % \end{cases} \\
    % W_{hh} &= T^{-1} \mathbf{s}
\end{align*}
The matrix $W_{ih}$ ``selects'' one part of the 
state encoding computed by $h'=W_{hh} \pi(q_i,e_j)$
via adding  $W_{ih} x_t$ to $h'$, where $\rho(e_t)=x_t$
is the encoded input at time step $t$. 
This addition makes the intermediate vector 
$W_{hh} \pi(q_i,e_j) +  W_{ih} x_t$ negative everywhere except
at the indexes corresponding to the current input. 

Additionally adding the bias term $b_{hh}$ ensures 
that the final vector computed 
before applying $\tanh$ is in a proper range and negative everywhere
 except at exactly one index. This term and the weights in $W_{ih}$ are
 multiplied by the saturation 
 factor $H_r$, such that after applying $\tanh$, we get
 approximately the encoding of a transition.
 
 For the output layer, we create a $2
 \times h$ matrix $W_{oh}$ following a similar approach as for $W_{hh}$. Let $\tau$ 
 be the one-hot encoding of $\mathit{true}$ and $\mathit{false}$. We
 require for each transition $(q_i, e_j)$ that
 $\sigma(W_{oh} \pi(q_i,e_j)) = \tau(\delta(q_i,e_j) \in F)$, 
 as we use the sigmoid function $\sigma$ as activation. In order 
 to use $\sigma$ in saturation,
 we again use a saturation factor $H_o$ and formulate
 this as an inequality  $W_{oh} \pi(q_i,e_j) \geq 
 H_o \tau(\delta(q_i,e_j) \in F)$. Finally, we solve a linear inequation system of $2\cdot |Q|\cdot |\Sigma|$ inequations to compute $W_{oh}$.
 
 Training an RNN with weights computed as described 
 above would yield a perfect RNN, where the state space
 would form dense clusters. However, there would be 
 $|Q| \cdot |\Sigma|$ many clusters, so we need to 
 analyze whether RNNs learn such non-minimal 
 representations. Moreover, attempting to further train
 such an RNN would usually stop immediately.
 For this reason, we add noise to the 
 weights $W_{ih}$, $W_{hh}$, $b_{ih}$, and $b_{hh}$
 to change the state transitions, so that we can analyze the effect
 of training \emph{close-to-perfect} RNNs. We use Gaussian noise with a mean 
 of $0$ and standard deviation of $wn$.

 \subsection{Appendix B -- Additional Evaluation Details}
 In the following, we provide additional details on the experimental
 setup and additional experimental results in tables.
 
 \subsubsection{Case Study Subjects.}
 We used regular language from the literature including three of the Tomita grammars~\cite{tomita:cogsci82}, where we have chosen Tomita 3, 5, and 7, as they define the most
 interesting languages. For example, Tomita 5 defines a parity
 language, which are difficult to learn for RNNs~\cite{DBLP:journals/tnn/GoudreauGCC94}. The Tomita grammars
 have five, four, and four states, respectively, and an alphabet of 
 size two. The MQTT server that we used in the evaluation has seven
 states, six inputs, and six outputs, and the regular expression 
 from Michaelenko et al.~\cite{DBLP:conf/iclr/MichalenkoSVBCP19} 
 has seven states and an alphabet containing four symbols.
 
 The randomly generated DFAs have five or ten states, 
 and alphabet sizes of two, four, or six.
For each combination of state and alphabet size, we generated five DFAs. We generated three Moore machines for each combination of eight and twelve states, two and four inputs, and three and five outputs.
Note that Moore machines encode regular languages that contain all possible input-output sequences. 
From an RNN perspective, they enable us to analyze how multi-class classification tasks affect clustering. 

As noted in Sect.~\ref{sec:experiments}, we constructed RNNs
to encode each of the DFAs. 
We applied the following parameters for saturation and noise in the construction of these close-to-perfect RNNs:
 \begin{compactitem}
 \item $H_r = 1.5$, $H_o = 1.5$, $wn = 0.05$
 \item $H_r = 2$, $H_o = 2$, $wn = 0.1$
 \item $H_r = 3$, $H_o = 3$, $wn = 0.2$
 \end{compactitem}

% \usepackage{multirow}
% \usepackage{graphicx}
\begin{table}[ht!]
\caption{Clustering ambiguity results from 1350 RNN's that achieved 80\% accuracy.}
\rowcolors{1}{}{black!20}
\resizebox{\textwidth}{!}{%
\begin{tabular}{|l|rr|rr|rr|r|}
\hline
\multicolumn{1}{|c|}{\multirow{2}{*}{\textbf{Clustering Function}}} &
  \multicolumn{2}{c|}{\textbf{Ambiguity}} &
  \multicolumn{2}{c|}{\textbf{\begin{tabular}[c]{@{}c@{}}Weighted\\ Ambiguity\end{tabular}}} &
  \multicolumn{2}{c|}{\textbf{\begin{tabular}[c]{@{}c@{}}Number of\\ Clusters\end{tabular}}} &
  \multicolumn{1}{c|}{\multirow{2}{*}{\textbf{\begin{tabular}[c]{@{}c@{}}\# Perfect\\ Clustering\end{tabular}}}} \\ \cline{2-7}
\multicolumn{1}{|c|}{} &
  \multicolumn{1}{c|}{Mean $\pm$ Std Dev} &
  \multicolumn{1}{c|}{Max} &
  \multicolumn{1}{c|}{Mean $\pm$ Std Dev} &
  \multicolumn{1}{c|}{Max} &
  \multicolumn{1}{c|}{Mean $\pm$ Std Dev} &
  \multicolumn{1}{c|}{Max} &
  \multicolumn{1}{c|}{} \\ \hline
LDA                 & \multicolumn{1}{r|}{0.01 $\pm$ 0.06}    & 0.933 & \multicolumn{1}{r|}{0.009 $\pm$ 0.05}   & 0.933 & \multicolumn{1}{r|}{8 + 2.5-}    & 15   & 1003 \\ \hline
Logistic Regression & \multicolumn{1}{r|}{0.003 $\pm$ 0.036}  & 0.935 & \multicolumn{1}{r|}{0.004 $\pm$ 0.039}  & 0.935 & \multicolumn{1}{r|}{8 $\pm$ 2.5}    & 12   & 1235 \\ \hline
DBSCAN   ($\epsilon=0.5$)           & \multicolumn{1}{r|}{0.002 $\pm$ 0.002}  & 0.025 & \multicolumn{1}{r|}{0.179 $\pm$ 0.239}  & 0.721 & \multicolumn{1}{r|}{200 $\pm$ 140}  & 531  & 368  \\ \hline
DBSCAN   ($\epsilon=0.25$)        & \multicolumn{1}{r|}{0.003 $\pm$ 0.002}  & 0.02  & \multicolumn{1}{r|}{0.29 $\pm$ 0.276}   & 0.772 & \multicolumn{1}{r|}{210 $\pm$ 118}  & 506  & 185  \\ \hline
DBSCAN   ($\epsilon=1$)       & \multicolumn{1}{r|}{0.005 $\pm$ 0.025}  & 0.672 & \multicolumn{1}{r|}{0.066 $\pm$ 0.15}   & 0.716 & \multicolumn{1}{r|}{143 $\pm$ 135}  & 568  & 591  \\ \hline
DBSCAN   ($\epsilon=1.5$)          & \multicolumn{1}{r|}{0.016 $\pm$ 0.063}  & 0.939 & \multicolumn{1}{r|}{0.05 $\pm$ 0.123}   & 0.939 & \multicolumn{1}{r|}{99 $\pm$ 115}   & 596  & 639  \\ \hline
DBSCAN   ($\epsilon=2$)         & \multicolumn{1}{r|}{0057 $\pm$ 0.144}   & 0.989 & \multicolumn{1}{r|}{0.110 $\pm$ 0.208}  & 0.98  & \multicolumn{1}{r|}{69 $\pm$ 90}    & 560  & 514  \\ \hline
k-means ($k=n$)            & \multicolumn{1}{r|}{0.359 $\pm$ 0.199}  & 0.863 & \multicolumn{1}{r|}{0.395 $\pm$ 0.207}  & 0.929 & \multicolumn{1}{r|}{8.5 $\pm$ 2.5}  & 12   & 49   \\ \hline
k-means ($k=n + 1$)         & \multicolumn{1}{r|}{0.322 $\pm$ 0.131}  & 0.811 & \multicolumn{1}{r|}{0.356 $\pm$ 0.201}  & 0.928 & \multicolumn{1}{r|}{9.5 $\pm$ 2.5}  & 13   & 68   \\ \hline
k-means ($k=n - 1$)         & \multicolumn{1}{r|}{0.401 $\pm$ 0.2}    & 0.895 & \multicolumn{1}{r|}{0.428 $\pm$ 0.21}   & 0.934 & \multicolumn{1}{r|}{7.5 - 2.5}   & 11   & 42   \\ \hline
k-means ($k=2n$)         & \multicolumn{1}{r|}{0.175 $\pm$ 0.154}  & 0.736 & \multicolumn{1}{r|}{0.2 $\pm$ 0.172}    & 0.92  & \multicolumn{1}{r|}{16 $\pm$ 5}     & 24   & 209  \\ \hline
k-means ($k=4n$)        & \multicolumn{1}{r|}{0.06 $\pm$ 0.091}   & 0.652 & \multicolumn{1}{r|}{0.068 $\pm$ 0.105}  & 0.904 & \multicolumn{1}{r|}{33 $\pm$ 10}    & 48   & 432  \\ \hline
k-means ($k=6n$)        & \multicolumn{1}{r|}{0.0291 $\pm$ 0.065} & 0.585 & \multicolumn{1}{r|}{0.032 $\pm$ 0.075}  & 0.878 & \multicolumn{1}{r|}{50 $\pm$ 15}    & 72   & 629  \\ \hline
k-means ($k=8n$)       & \multicolumn{1}{r|}{0.017 $\pm$ 0.051}  & 0.54  & \multicolumn{1}{r|}{0.0181 $\pm$ 0.057} & 0.674 & \multicolumn{1}{r|}{67 $\pm$ 21}    & 96   & 783  \\ \hline
OPTICS              & \multicolumn{1}{r|}{0.006 $\pm$ 0.003}  & 0.06  & \multicolumn{1}{r|}{0.134 $\pm$ 0.071}  & 0.544 & \multicolumn{1}{r|}{156 $\pm$ 20}   & 206  & 16   \\ \hline
mean shift  ($bw=\alpha$)        & \multicolumn{1}{r|}{0.847 $\pm$ 0.244}  & 0.998 & \multicolumn{1}{r|}{0.878 $\pm$ 0.201}  & 0.998 & \multicolumn{1}{r|}{1.35 $\pm$ 1.2} & 20   & 84   \\ \hline
mean shift  ($bw=\alpha/2$)      & \multicolumn{1}{r|}{0.038 $\pm$ 0.063}  & 0.935 & \multicolumn{1}{r|}{0.073 $\pm$ 0.136}  & 0.958 & \multicolumn{1}{r|}{37 $\pm$ 51}    & 829  & 373  \\ \hline
mean shift  ($bw=\alpha/4$)     & \multicolumn{1}{r|}{0.001 $\pm$ 0.014}  & 0.246 & \multicolumn{1}{r|}{0.011 $\pm$ 0.086}  & 0.943 & \multicolumn{1}{r|}{136 $\pm$ 178}  & 1425 & 1296 \\ \hline
mean shift  ($bw=\alpha/8$)      & \multicolumn{1}{r|}{0.0004 $\pm$ 0.005} & 0.09  & \multicolumn{1}{r|}{0.005 $\pm$ 0.06}   & 0.912 & \multicolumn{1}{r|}{391 $\pm$ 397}  & 1448 & 1331 \\ \hline
\end{tabular}%
}
\label{tab:normal_training_table}
\end{table}

% \usepackage{multirow}
% \usepackage{graphicx}
\begin{table}[ht!]
\caption{Results from 94 RNNs that were trained to 100\% accuracy after introducing noise to ``correct-by-construction'' RNN.}
\rowcolors{1}{}{black!20}
\resizebox{\textwidth}{!}{%
{\scriptsize
\begin{tabular}{|l|rr|rr|rr|r|}
\hline
\multicolumn{1}{|c|}{\multirow{2}{*}{\textbf{Clustering Function}}} & \multicolumn{2}{c|}{\textbf{Ambiguity}}                              & \multicolumn{2}{c|}{\textbf{\begin{tabular}[c]{@{}c@{}}Weighted\\ Ambiguity\end{tabular}}} & \multicolumn{2}{c|}{\textbf{\begin{tabular}[c]{@{}c@{}}Number of\\ Clusters\end{tabular}}} & \multicolumn{1}{c|}{\multirow{1}{*}{\textbf{\begin{tabular}[c]{@{}c@{}}\# Perfect\\ Clustering \end{tabular}}}} \\ \cline{2-7}
\multicolumn{1}{|c|}{}                                              & \multicolumn{1}{c|}{Mean $\pm$ Std Dev}      & \multicolumn{1}{c|}{Max} & \multicolumn{1}{c|}{Mean $\pm$ Std Dev}                 & \multicolumn{1}{c|}{Max}            & \multicolumn{1}{c|}{Mean $\pm$ Std Dev}               & \multicolumn{1}{c|}{Max}              & \multicolumn{1}{c|}{}                                                                                          \\ \hline
LDA                                                                 & \multicolumn{1}{r|}{0.0007 $\pm$ 0.006}      & 0.06                     & \multicolumn{1}{r|}{0.0004 $\pm$ 0.004}                 & 0.039                               & \multicolumn{1}{r|}{7 $\pm$ 2.5}                      & 10                                    & 88                                                                                                             \\ \hline
Logistic Regression                                                 & \multicolumn{1}{r|}{8.22e-06 $\pm$ 5.8e-05}  & 0.0005                   & \multicolumn{1}{r|}{9.01e-06 $\pm$ 6.3e-05}             & 0.0005                              & \multicolumn{1}{r|}{7 $\pm$ 2.5}                      & 10                                    & 92                                                                                                             \\ \hline
DBSCAN ($\epsilon=0.5$)                                                              & \multicolumn{1}{r|}{0.002 $\pm$ 0.009}       & 0.07                     & \multicolumn{1}{r|}{0.001 $\pm$ 0.014}                  & 0.14                                & \multicolumn{1}{r|}{24.5 $\pm$ 15.8}                  & 61                                    & 80                                                                                                             \\ \hline
DBSCAN ($\epsilon=0.25$)                                                      & \multicolumn{1}{r|}{0.007 $\pm$ 0.009}       & 0.03                     & \multicolumn{1}{r|}{0.0004 $\pm$ 0.0008}                & 0.005                               & \multicolumn{1}{r|}{31 $\pm$ 20.5}                    & 100                                   & 51                                                                                                             \\ \hline
DBSCAN ($\epsilon=1$)                                                         & \multicolumn{1}{r|}{0.005 $\pm$ 0.018}       & 0.12                     & \multicolumn{1}{r|}{0.018 $\pm$ 0.069}                  & 0.492                               & \multicolumn{1}{r|}{23 $\pm$ 16}                      & 60                                    & 82                                                                                                             \\ \hline
DBSCAN ($\epsilon=1.5$)                                                      & \multicolumn{1}{r|}{0.478 $\pm$ 0.112}       & 0.9                      & \multicolumn{1}{r|}{0.0225 $\pm$ 0.212}                 & 0.907                               & \multicolumn{1}{r|}{20 $\pm$ 15}                      & 56                                    & 42                                                                                                             \\ \hline
DBSCAN ($\epsilon=2$)                                                        & \multicolumn{1}{r|}{0.412 $\pm$ 0.348}       & 0.99                     & \multicolumn{1}{r|}{0.55 $\pm$0.35}                     & 0.996                               & \multicolumn{1}{r|}{7 $\pm$ 7.5}                      & 34                                    & 10                                                                                                             \\ \hline
k-means ($k=n$)                                                    & \multicolumn{1}{r|}{0.376 $\pm$ 0.203}       & 0.846                    & \multicolumn{1}{r|}{0.429 $\pm$ 0.214}                  & 0.863                               & \multicolumn{1}{r|}{7 $\pm$ 2.5}                      & 10                                    & 4                                                                                                              \\ \hline
k-means ($k=n+1$)                                                           & \multicolumn{1}{r|}{0.320 $\pm$ 0.183}       & 0.758                    & \multicolumn{1}{r|}{0.369 $\pm$ 0.203}                   & 0.774                               & \multicolumn{1}{r|}{8 $\pm$ 2.5}                      & 11                                    & 3                                                                                                              \\ \hline
k-means ($k=n-1$)                                                           & \multicolumn{1}{r|}{0.431 $\pm$ 0.210}       & 0.88                     & \multicolumn{1}{r|}{0.491 $\pm$ 0.219}                  & 0.892                               & \multicolumn{1}{r|}{6 $\pm$ 2.5}                      & 9                                     & 4                                                                                                              \\ \hline
k-means ($k=2n$)                                                           & \multicolumn{1}{r|}{0.129 $\pm$ 0.130}       & 0.43                     & \multicolumn{1}{r|}{0.166 $\pm$ 0.175}                  & 0.524                               & \multicolumn{1}{r|}{14 $\pm$ 5}                       & 20                                    & 38                                                                                                             \\ \hline
k-means ($k=4n$)                                                          & \multicolumn{1}{r|}{0.019 $\pm$ 0.045}       & 0.136                    & \multicolumn{1}{r|}{0.029 $\pm$ 0.06}                   & 0.208                               & \multicolumn{1}{r|}{28 $\pm$ 10}                      & 40                                    & 71                                                                                                             \\ \hline
k-means ($k=6n$)                                                           & \multicolumn{1}{r|}{4.1e-05 $\pm$ 0.0002}    & 0.001                    & \multicolumn{1}{r|}{2.24e-05 $\pm$ 0.0001}              & 0.0006                              & \multicolumn{1}{r|}{42 $\pm$ 15}                      & 60                                    & 90                                                                                                             \\ \hline
k-means ($k=8n$)                                                           & \multicolumn{1}{r|}{2.1e-05 $\pm$ 0.0001}    & 0.0006                   & \multicolumn{1}{r|}{1.854e-05 $\pm$ 9.4e-05}            & 0.0006                              & \multicolumn{1}{r|}{55 $\pm$ 19}                      & 80                                    & 90                                                                                                             \\ \hline
OPTICS                                                              & \multicolumn{1}{r|}{0.003 $\pm$ 0.0008}      & 0.004                    & \multicolumn{1}{r|}{0.106 $\pm$ 0.03}                   & 0.259                               & \multicolumn{1}{r|}{270 $\pm$ 35}                     & 336                                   & 3                                                                                                              \\ \hline
mean shift  ($bw=\alpha$)                                                         & \multicolumn{1}{r|}{0.871 $\pm$ 0.253}       & 0.998                    & \multicolumn{1}{r|}{0.881 $\pm$ 0.239}                  & 0.998                               & \multicolumn{1}{r|}{1.3 $\pm$ 1.3}                    & 9                                     & 5                                                                                                              \\ \hline
mean shift  ($bw=\alpha/2$)                                                     & \multicolumn{1}{r|}{0.001 $\pm$ 0.006}       & 0.045                    & \multicolumn{1}{r|}{0.002 $\pm$ 0.008}                  & 0.058                               & \multicolumn{1}{r|}{24 $\pm$ 15}                      & 60                                    & 85                                                                                                             \\ \hline
mean shift  ($bw=\alpha/4$)                                                      & \multicolumn{1}{r|}{8.07e-06 $\pm$ 7.7e-0.5} & 0.0007                   & \multicolumn{1}{r|}{1.5e-05 $\pm$ 0.0001}               & 0.001                               & \multicolumn{1}{r|}{25 $\pm$ 15}                      & 62                                    & 93                                                                                                             \\ \hline
mean shift ($bw=\alpha/8$)
& \multicolumn{1}{r|}{0}                    & 0                        & \multicolumn{1}{r|}{0}                               & 0                                   & \multicolumn{1}{r|}{36 $\pm$ 28}                      & 152                                   & 94                                                                                                             \\ \hline
\end{tabular}%
}
}
\label{tab:retrained}
\end{table}
\subsubsection{Additional Results.}
In Table~\ref{tab:normal_training_table} and Table~\ref{tab:retrained}, we show detailed results from our clustering experiments performed on RNNs trained from scratch and on RNNs initialized with close-to-perfect weights, respectively. 
The tables
show similar trends as the figures in the main part of the paper, but 
provide more details. 

\subsubsection{Extraction of cluster automata.}

\begin{figure}[ht!]
    \centering
    \includegraphics[width=\textwidth]{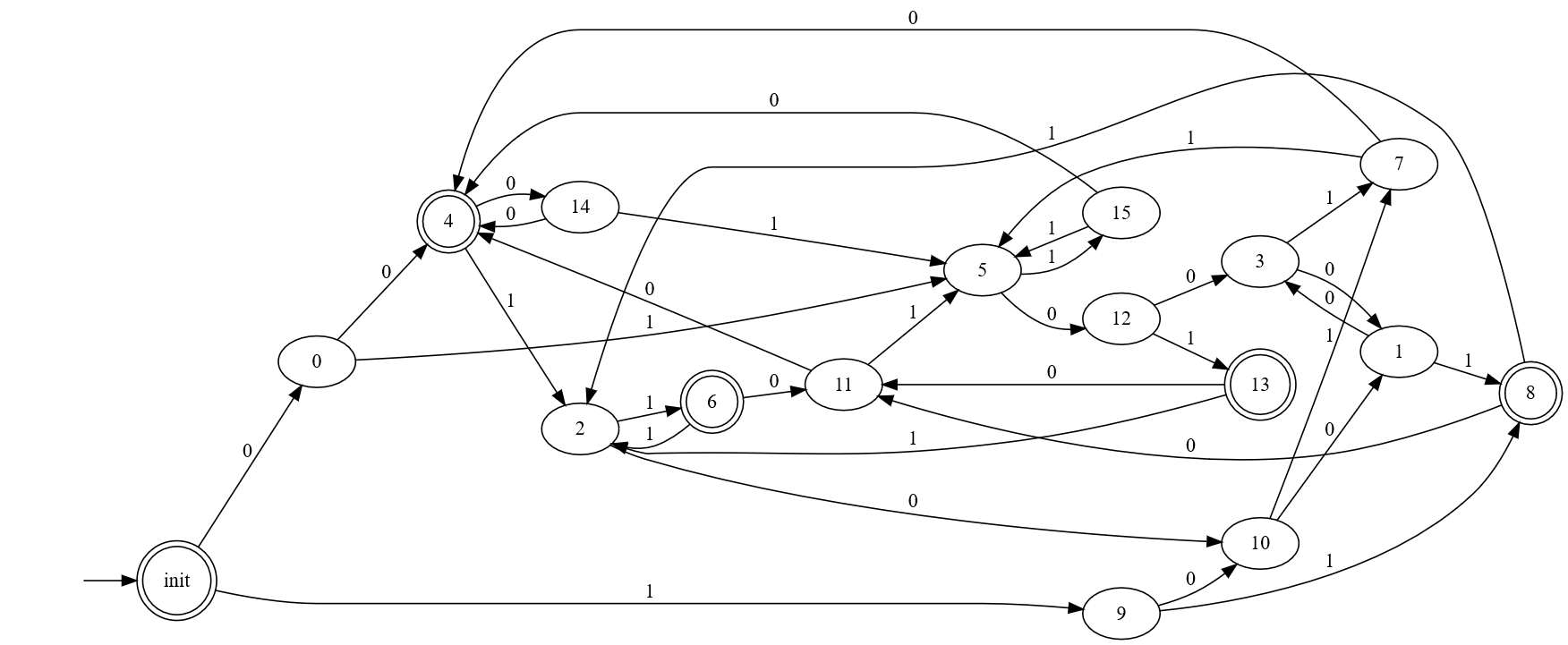}
    \caption{Extracted cluster automata from an RNN trained on the Tomita 5 grammar. K-means clustering  with $k=4n$ was used. The extracted model is a non-minimal representation of the ground truth model shown in Fig.~\ref{fig:extracted_cluster_automaton}.}
\label{fig:extracted_cluster_automaton}
\end{figure}

Figure~\ref{fig:extracted_cluster_automaton} depicts a non-minimal automaton extracted from an RNN trained on the Tomita 5 grammar. The extracted model is a non-minimal representation of the ground truth model, which is shown on the left-hand side of Fig.~\ref{fig:training_and_extraction}. We have applied the extraction method found in~\cite{DBLP:journals/tnn/HouZ20} with multiple clustering functions that achieved perfect accuracy, where Fig.~\ref{fig:extracted_cluster_automaton} is based on k-means clustering
with $k=4n$. Basically, we create one DFA state for each cluster $k\in K$. 
For the transitions, suppose that $(h,q)$ and $(h',q')$ are pairs of hidden states and automaton states collected consecutively while processing symbol $e$.
In this case, we add a transitions from $c(h)$ to $c(h')$ labeled by $e$,
where $c$ is the clustering function. If the RNN outputs $\mathit{true}$
for one hidden state in a cluster $k$, we add $k$ to the accepting states.
While the extracted model size varied, all extracted automata were non-minimal representations of the ground-truth model that was used to train the RNN. Our findings conform to those of~\cite{DBLP:journals/tnn/HouZ20} and they indicate that RNNs learn non-minimal representations of regular languages.

\end{document}